\documentclass[]{LeafLabLaTexTemplate}
\input{preamble}
\usepackage{wrapfig}
    
\newcolumntype{Y}{>{\centering\arraybackslash}X}
\makeatletter

\newcommand{\barebonestoc}{\@starttoc{toc}}
\makeatother

\title{\textcolor{leafblue}{OccSim}: Multi-kilometer Simulation with Long-horizon Occupancy World Models}

\author[]{Tianran Liu$^\star$}
\author[]{Shengwen Zhao}
\author[]{Mozhgan Pourkeshavarz}
\author[]{Weican Li}
\author[]{Nicholas Rhinehart$^\dagger$}

\affiliation[]{Learning, Embodied Autonomy, and Forecasting (LEAF) Lab, University of Toronto}

\abstract{
Data-driven autonomous driving simulation has long been constrained by its heavy reliance on pre-recorded driving logs or spatial priors, such as HD maps. This fundamental dependency severely limits scalability, restricting open-ended generation capabilities to the finite scale of existing collected datasets. To break this bottleneck, we present OccSim, the first 3D occupancy world model driven simulator that requires neither HD maps nor log-derived layouts. OccSim obviates the requirement for continuous logs or HD maps; conditioned only on a single initial frame and a sequence of future ego-actions. It can stably generate over 3000 continuous frames, an $>\!\!80\times$ improvement in stable generation length over previous SoTA occupancy world models, and fuses them into maps spanning over 9 kilometers. Further qualitative stress tests show that it can roll out beyond 110K frames. OccSim is powered by two modules: W-DiT based static occupancy world model and the Layout Generator. W-DiT handles the ultra-long-horizon generation of static environments by explicitly introducing known rigid transformations in architecture design, while the Layout Generator populates the dynamic foreground with reactive agents based on the synthesized road topology. With these designs, OccSim can synthesize massive, diverse simulation streams. Extensive experiments demonstrate its downstream utility: data collected directly from OccSim can pre-train 4D semantic occupancy forecasting models to achieve up to 67\% zero-shot performance on unseen data, outperforming previous asset-based simulator by 11\%. When the collected data is scaled up to  $5\times$ its original size, zero-shot performance increases to 74\%.
}

\metadata[
\raisebox{-0.35em}{\includegraphics[width=0.05\linewidth]{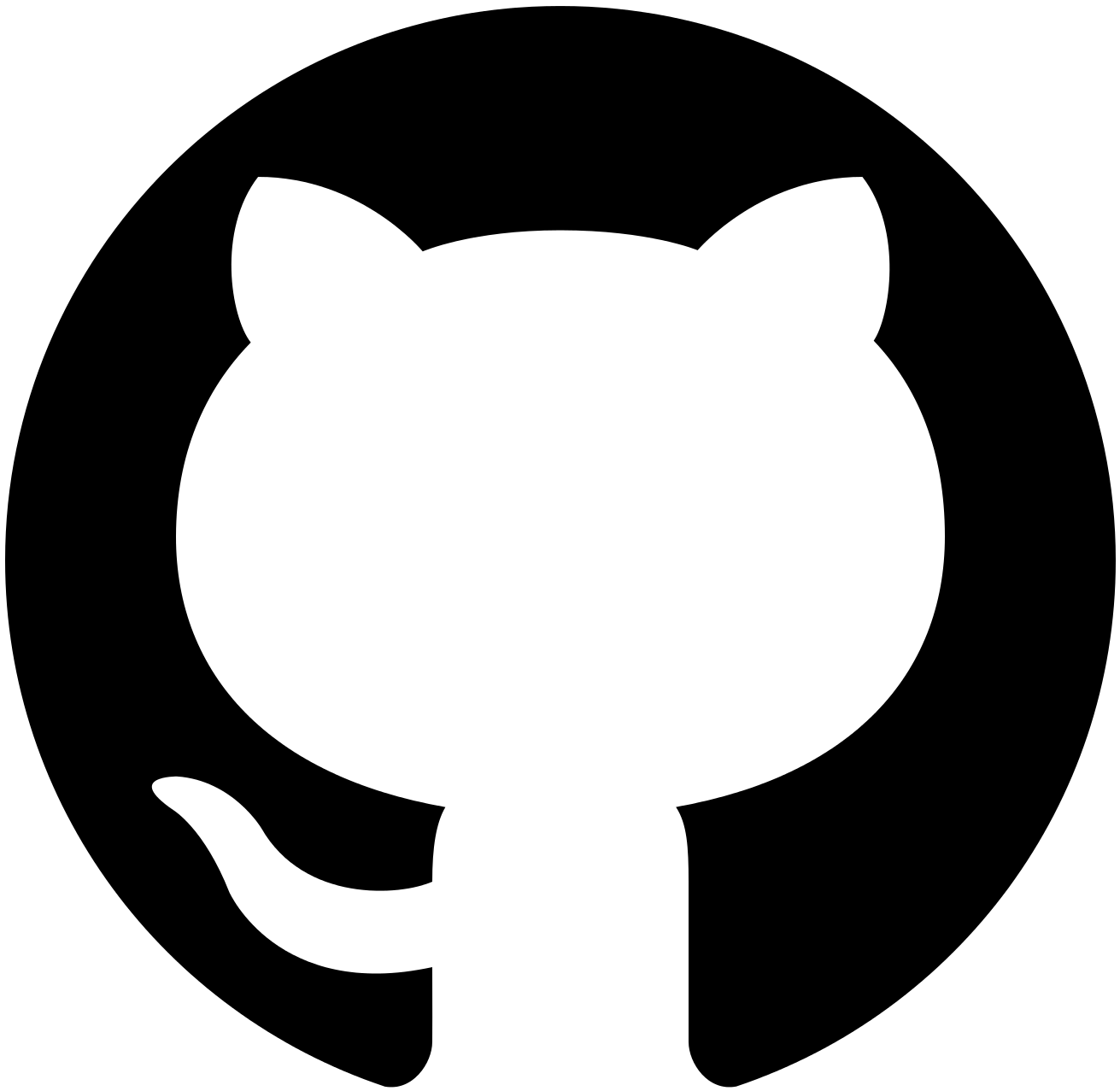}}~~ Project Website]{\href{https://orbis36.github.io/OccSim/} {\textbf{https://orbis36.github.io/OccSim/}}
}

\begin{document}

\maketitle

\begin{figure}[t]
    \centering
    \includegraphics[width=\linewidth]{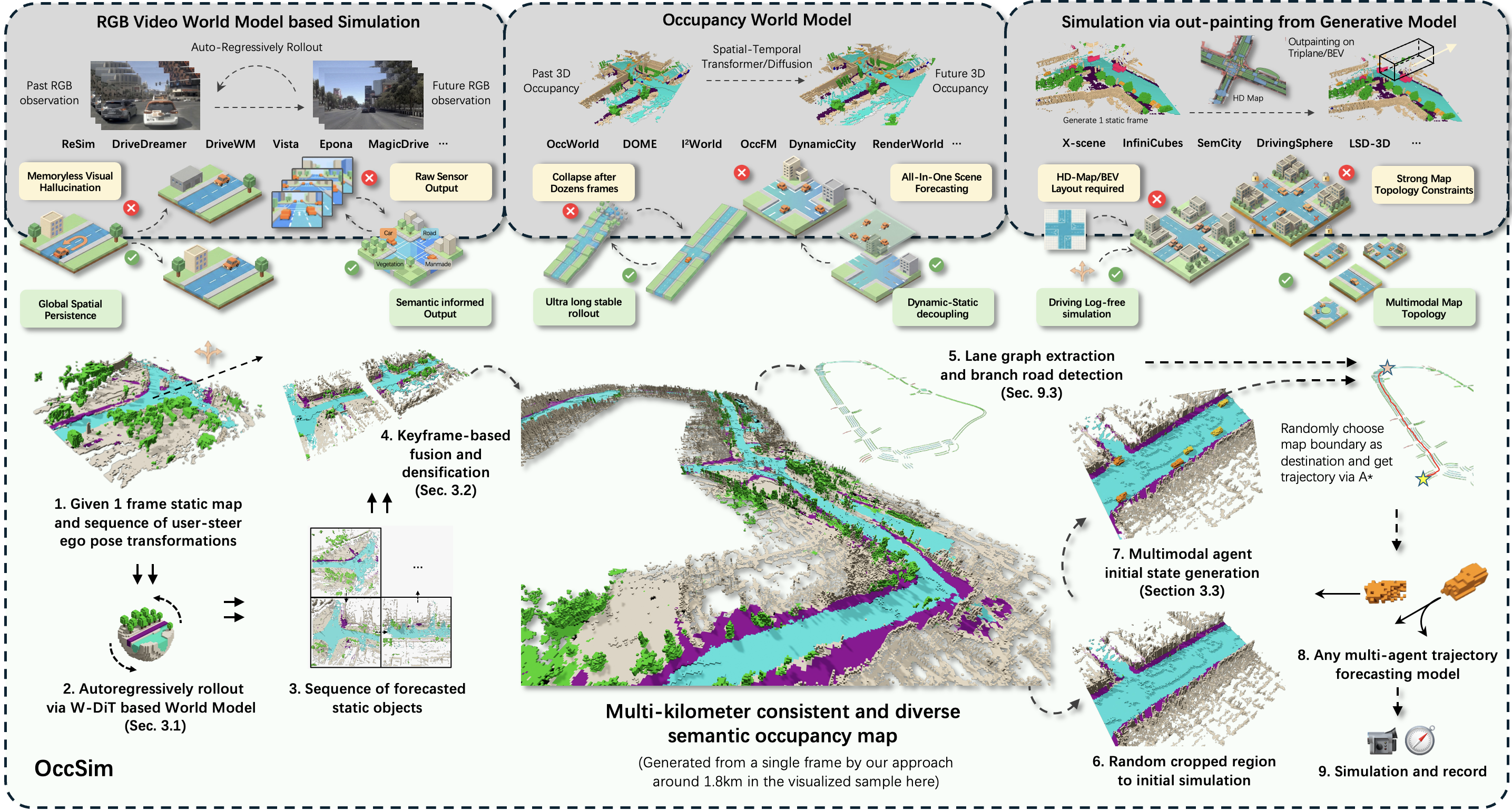}
    \vspace{-0.6cm}
    \caption{Comparison of OccSim and workflow. Given a single frame and a sequence of future actions, the proposed W-DiT builds a multi-kilometer consistent occupancy map for autonomous driving simulation, without HD maps or driving logs at inference time. The agent initial poses are generated via the layout generator and then forward-simulated. OccSim is compatible with multi-agent forward simulation methods as plug-and-play modules; in our experiments, we use an IDM variant \cite{kesting2010enhanced}. }
    \label{fig:main_figure}
\end{figure}

\section{Introduction}
\label{sec:intro}

Simulation is indispensable for the advancement of Level 4 autonomous driving systems \cite{caesar2021nuplan, dauner2024navsim, codevilla2018offline}. Existing work basically obtain these simulation environments in one of two ways, and both are capped by human effort: Replay of recorded data provides authentic data streams, it inherently lacks reactive responses to ego behaviors. In contrast, geometry-based simulators (e.g., CARLA\cite{Dosovitskiy17}), despite supporting interactivity, suffer from a significant sim-to-real gap regarding sensor realism and content diversity \cite{yang2023unisim, zhang2025epona}. This creates a fundamental trade-off: current approaches are often forced to compromise between realism, diversity, and interactivity. Although emerging data-driven simulations \cite{yan2025drivingsphere, zhou2024hugsim} promise to overcome this dilemma, they remain conditioned on HD maps or BEV layouts, so the set of environments they can produce is still bounded by the coverage of annotated maps rather than by the model itself. \blfootnote{$^*$: Contact: tianran.liu@mail.utoronto.ca.  $\quad$  $\dagger$: Project Manager.}

Specifically, we argue that a comprehensive and scalable data-driven simulator must satisfy the following desiderata: 1. \textbf{Sensor realism \& semantic richness}: The simulator should not only approximate the distribution of real-world sensor data (high-fidelity sensor realism) to provide diverse testbed, but also should inherently generate dense semantic labels, eliminating the reliance on manual annotation. 2. \textbf{Controllable \& plausible agents}: infer initial traffic states from the environment rather than replaying logs, and expose agent control as a replaceable interface accepting anything from microscopic heuristics to learned forecasters. 3. \textbf{Global spatial persistence}: The simulator should maintain global spatial persistence, overcoming the ``memoryless visual hallucination'' prevalent in standard generative models. During long-horizon  driving, static elements (e.g., road topology, buildings) must preserve strict spatiotemporal consistency. For instance, when the ego-vehicle completes a loop and revisits a previously observed location, the static environment must remain structurally identical. 4. \textbf{Log-independent Simulation}: The simulator's operation should not be overly-constrained by pre-recorded driving logs. Ideally, it possesses the capability for open-ended generation in novel, unseen scenarios in order to overcome the scale limitations of collected data.

In this paper, we aim to address the overarching question: \emph{``How can we construct a data-driven generative simulator that efficiently satisfies these desiderata?''} To this end, we present \textbf{OccSim}. To the best of our knowledge, OccSim is the first system to drive driving simulation from an occupancy world model, generating multi-kilometer, globally consistent environments without any HD map or BEV layout as input. As illustrated in Figure \ref{fig:main_figure}, it satisfies the above desiderata by conditioning solely on a single static frame and a sequence of future ego-actions, while being trained entirely on publicly-available data.

Specifically, we adopt a decoupling strategy for static and dynamic voxels, modelling simulation as two synergistic generative tasks: a multi-kilometer rollout of the world model that assembles the large-scale static scene, and the conditional generation of dynamic traffic flow within it. These two tasks are also decoupled in time: the static world is generated once and then frozen, whereas agents are controlled online on top of it. For static world generation, the core difficulty is sustaining stable, high-quality and diverse output over long horizons. Prior work lacks this stability: given future actions, SOTA occupancy world models \cite{shi2025come, liao2025i2} can only stably generate fewer than 50 frames (Fig. \ref{fig:uncond_2d}), while outpainting-based methods need human-labelled HD-Map \cite{lu2025infinicube} or trade stability for diversity \cite{lee2024semcity}: their fidelity  degrades sharply after certain times of rollouts and the resulting maps remain topologically fragmented (Fig. \ref{fig:comp_non_wm}).

Such a horizon suffices for vehicle control decisions, but not for large-scale map construction (e.g., multiple city blocks, $\geq\!2\mathrm{km}^2$) or long distance task simulation: at a 2Hz sampling rate with roughly 3 meters of ego motion per frame, 20--50 frames span only about 60--150m, far short of real street scale.  

We attribute this short rollout horizon to the architecture of current stochastic 4D occupancy world models, which largely inherit their design from video generation: multiple 3D occupancy latents are stacked along the temporal dimension, and features are then aggregated by DiTs over both temporal and spatial dimensions \cite{gu2024dome, ma2024latte}. This paradigm ignores the geometric rigidity of 3D space and forces the network to hallucinate viewpoint changes implicitly, even though most of the observed region at t+1 is related to t by a known rigid transformation. What could be supplied as an exact geometric prior is instead left to be learned, so the residual error is re-injected at every step and compounds over autoregressive rollout until the scene collapses.

Based on this observation, we propose ``W-DiT'', the first occupancy world model capable of diverse, at least 3k-step rollouts of diverse static maps under complex trajectories (tested with over 3,000 frames generated, >80$\times$ improvement in stable rollout length). The core mechanism of W-DiT is a pipeline designed for effective conditioning on the previous frame: W-DiT explicitly utilizes 3D geometric priors through the proposed Mask-injected Conditioning mechanism. Beyond these, we also provide an SNR-weighted occupancy perception loss further supervises the prediction in decoded occupancy space, which we find necessary to keep rollout quality stable over thousands of frames. Orthogonally, we show that a norm-invariant spherical semantic readout in the occupancy VAE reshapes the latent manifold in a way that any occupancy world model can benefit from.

Single-frame outputs are then consolidated into a persistent static map: we design a keyframe-based fusion scheme and a graph-search-based lane topology extraction algorithm, yielding a large-scale road network that agents can drive on. Populating it without historical logs requires a generative treatment of traffic layout. Rather than placing agents by hand-tuned spawning rules, we train a layout generator based on latent flow matching to produce initial traffic states conditioned on the local road structure; agent motion is then rolled out by a replaceable control module, for which we use a 2D-IDM variant. The inference logic of OccSim is shown in Fig. \ref{fig:main_figure}: generate the large-scale static occupancy, fuse them into a large map, extract lanes graph, sample an ego position within it, then populate and control all agents.
\vspace{1em}
\begin{enumerate}

    \item \textbf{A simulation pipeline that requires no logs or maps as input.} We propose a new pipeline for data-driven occupancy-based simulation: an occupancy world model generates the large-scale static scene, a layout generator infers initial agent states directly from the generated semantic occupancy, and agent motion is produced by a replaceable control module, instantiated here with a 2D-IDM variant, and compatible with any modern trajectory forecasting model. 
    
    \item \textbf{An occupancy world model stable over ultra-long horizons.} Using the same  training data, our W-DiT-based backbone optimized with the proposed objective achieves a stable rollout length over $\mathbf{80\times}$ longer than before. This makes W-DiT, to our knowledge, the first model to roll out over \textbf{3,000 frames} of static occupancy and form complex road topologies without relying on HD maps in any form. Under user-steered trajectories, further stress tests qualitatively show stable and diverse rollout beyond \textbf{110K frames} (Fig.~\ref{fig:traj_100km}), building a 100km$+$ city-scale occupancy map.

    \item \textbf{An evaluation protocol for log-free environment generation.} Beyond the logged horizon no paired ground truth exists, and each individual metric admits a trivial optimum. We therefore report fidelity, diversity and road-network connectivity jointly at matched spatial scales, reference them against an explicit collapse baseline, and ground them in the sim-to-real transfer of a downstream forecasting task, an external judge the generator cannot optimise against. Under this protocol, forecasting models trained purely on OccSim data reach 67\% of in-domain IoU (vs. 60\% for CARLA). These numbers increase to 74\% on the $5\times$ data scale, 22.1\% better than CARLA.
    
\end{enumerate}

\section{Related Work}

\subsection{Long horizon RGB world models}

Driven by the abundance of data resources, numerous efforts to construct world models for autonomous driving have utilized RGB signals as their primary input to achieve robust long-horizon rollouts. For instance, GAIA-1 \cite{hu2023gaia} employs a multi-modal transformer architecture with language and action conditioning to autoregressively generate minute-long video clips. GAIA-2 \cite{russell2025gaia} extended this capability to multi-view RGB generation by transitioning the underlying backbone from a deterministic model to flow matching. Furthermore, Cosmos-Drive-Dreams \cite{ren2025cosmos} finetuning the Cosmos \cite{agarwal2025cosmos} foundation model to enhance the roll out of autonomous driving scene generation. In the indoor navigation task, Ego2Memory \cite{zhang2025mem2ego} explicitly uses VLM to map 2D features with Frontier Map for long-term persistent rollout.

By introducing a two-stage training and fine-tuning pipeline alongside better utilization of dynamic priors, Vista \cite{gao2024vista} significantly improves the resolution and realism of generated videos, as evidenced by FID and FVD metrics. Subsequently, DrivingWorld \cite{hu2024drivingworld} adopted a purely auto-regressive (AR) based approach. More recently, Epona \cite{zhang2025epona} provided the first open-source implementation for long video rollouts, utilizing an AR diffusion method. This AR-plus-diffusion paradigm has since been extended to unified world models that jointly perform understanding, planning, and generation within a single VLM backbone \cite{zhou2025hermes, xiong2026unidrive}.

\subsection{Occupancy World Models in Autonomous Driving}

While RGB images offer rich semantic features, accurately capturing and predicting the complex 3D geometry and physical depth of dynamic environments remains a significant challenge. Consequently, there is a growing trend towards developing world models based on 3D occupancy representations, which are typically derived from voxelized and densified LiDAR data.

Earlier works such as OccWorld \cite{zheng2024occworld} and Occllama \cite{wei2024occllama} addressed this by combining discrete encoding techniques (e.g., VQ-VAE \cite{van2017neural}) with deterministic auto-regressive (AR) transformer architectures, with Occllama further incorporating Large Language Models (LLMs) to enhance scene comprehension. Currently, generative modeling approaches have gained significant traction. For instance, DOME \cite{gu2024dome} utilizes a continuous VAE alongside a high-performance diffusion model, enabling fine-grained controllability. Recent advancements have further extended these capabilities: COME \cite{shi2025come} introduces a post-trained ControlNet \cite{zhang2023adding} approach, DynamicCity \cite{bian2024dynamiccity} leverages hex-plane decomposition, and OccFM \cite{liu2025towards} employs a Flow Matching-based multi-scale DiT \cite{peebles2023scalable} architecture. Expanding the scope, UniScene \cite{li2025uniscene} explores a conditional pipeline that uses BEV layouts to generate comprehensive multi-sensor data, ranging from RGB images to LiDAR point clouds. 

\subsection{Data driven simulation with generative models}

High-fidelity simulation is crucial for bridging the sim-to-real gap in autonomous driving, yet conventional physics-based simulators like CARLA \cite{Dosovitskiy17} often struggle with domain discrepancies. Recent research has shifted towards data-driven approaches to leverage large-scale real-world datasets. Nocturne \cite{vinitsky2022nocturne} facilitates high-efficiency simulation by replaying trajectories from the Waymo dataset. Building upon this, SLEDGE \cite{chitta2024sledge} and Scenario Dreamer\cite{rowe2025scenario} extends simulation to synthesized scenarios by generatively modeling road layouts and vehicle agents. In parallel, efforts have been made to generate more realistic sensory observations. ReSim \cite{yang2025resim} and DriveArena \cite{yang2025drivearena} utilize video generation techniques and HD maps to render high-resolution RGB observations for closed-loop policy evaluation. 

The raising of 3D Gaussian Splatting rendering in unbounded scene \cite{chen2024lidar, zhou2024hugsim, xiao2025splatco} allow us 
reconstruct static backgrounds more realistic. While, the SemCity\cite{lee2024semcity} and InifiCube\cite{lu2025infinicube} follow the out-painting solution introduced by Repainting\cite{lugmayr2022repaint} while combine it with tri-plane decomposition diffusion. Most recently, X-scene \cite{yang2025x} advances the field by proposing an external diffusion module that generates BEV layouts from natural language prompts, thus mitigating dependency on recorded driving logs. 

\section{Long-horizon static Occupancy World Models}

OccSim organizes simulation into two loops on different timescales. The \emph{generation loop} runs once per environment: W-DiT rolls out static occupancy along a given trajectory, and the frames are fused into a single global map, which is then frozen. The \emph{simulation loop} runs every timestep on top of it: the ego and all background agents re-plan online from their instantaneous neighbourhood states, querying the static environment rather than regenerating it.

Next, we will introduce the W-DiT architecture and its corresponding optimization methods in Section \ref{sec:w-dit}, which form the foundation of our infinite-horizon static scene generation. In Section \ref{sec:fusion}, we briefly describe how to merge single-frame static scenes generated by W-DiT and infinitely extend them into loop closing city-scale environments. Furthermore, in Section \ref{sec:agent}, we will demonstrate how to learn the possible distribution of agents layout from data and incorporate them into the scene to interact with test ego vehicle.

\subsection{Long-horizon stable generation with W-DiT}
\label{sec:w-dit}

\begin{figure}[tb]
    \centering
    \includegraphics[width=0.95\linewidth]{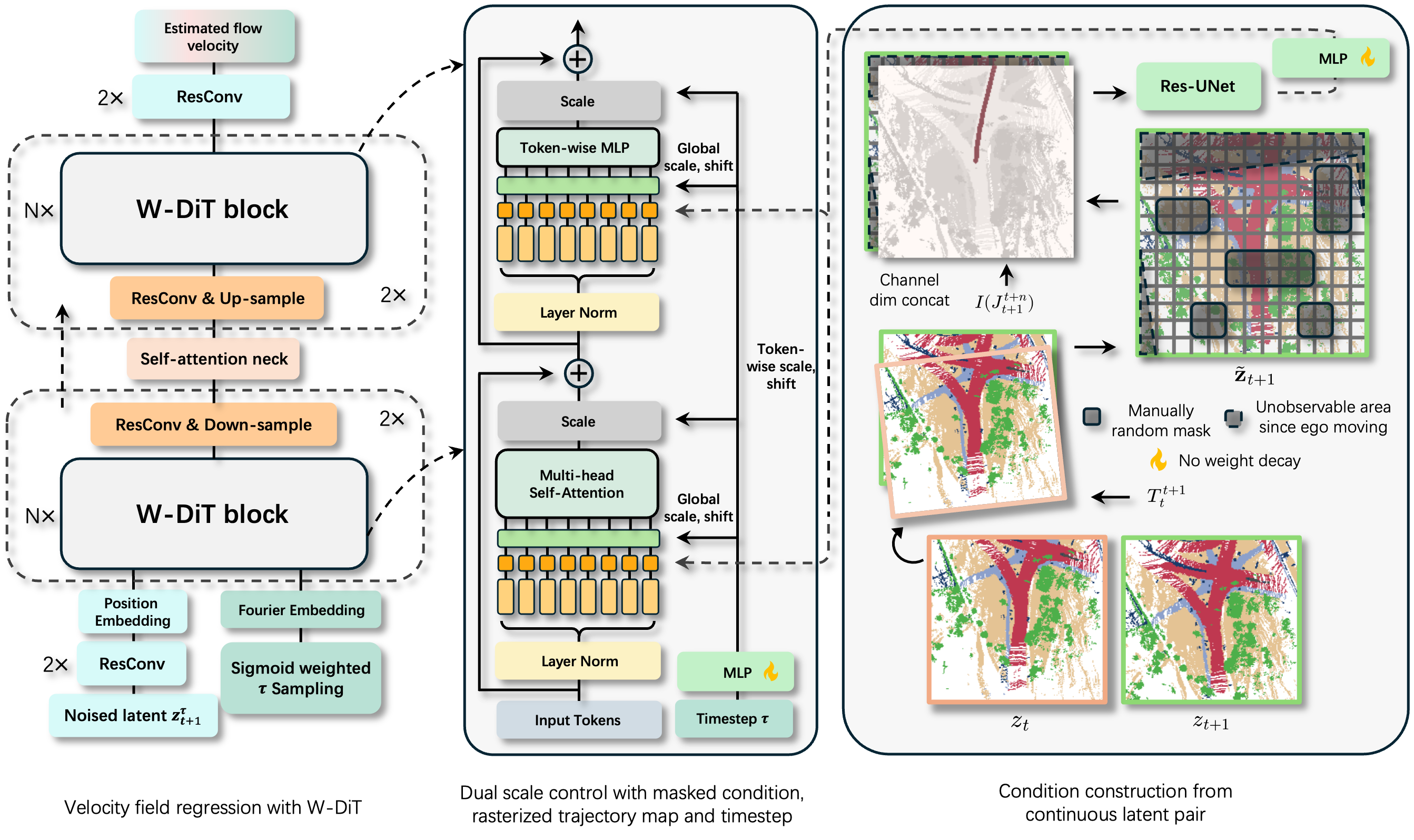}
    \caption{Illustration of structure to generate road map of $t+1$ from condition at timestep t. Here, $t$ represents the sequence frame index, and $\tau$ denotes the probability flow timestep.  This token-wise scale and shift condition injection method allows precise control of the rigid transformation during long-horizon generation. $I(\cdot)$ stands for rasterization process of given future trajectory $J_{t+1}^{t+n}$. The latent representations $z_t$ and $z_{t+1}$ ($\in \mathbb{R}^{H \times W \times C_z}$) at bottom right corner are visualized as decoded BEV occupancy maps for intuitive illustration.}
    \label{fig:W-DiT}
    \vspace{-1em}
\end{figure}

We find that a geometry-aware alternative to the standard 2+1D aggregation yields substantial improvements in long-term stability. Formally, using an occupancy VAE~\citep{liu2025towards} (consisting of spatial compressor $\mathcal{E}$ and decoder $\mathcal{D}$), we can map the occupancy frames $\mathcal{O}_t, \mathcal{O}_{t+1}$ to their respective latent representations $\mathbf{z}_t, \mathbf{z}_{t+1} \in \mathbb{R}^{H \times W \times C_z}$, where $t$ denotes time index. Our aim is to model the joint predictive distribution over an extended future horizon: given an initial latent state $\mathbf{z}_0$ and a sequence of future ego-trajectories $J_{0:K+\sigma}$, our goal is to auto-regressively estimate:
\vspace{-0.2em}
\begin{equation}
    P(\mathbf{z}_{1:K} | \mathbf{z}_0, J_{0:K+\sigma}) = \prod_{i=1}^{K} P_{\theta}(\mathbf{z}_{i} | \mathbf{z}_{i-1}, J_{i-1:i+\sigma-1}),
\end{equation}%
\vspace{-0.5em}

where $K$ is the forecast horizon and $K+\sigma$ the trajectory length. For $K \geq 10^2$, the per-step residual of a learned re-projection compounds into structural drift. The W-DiT block (Fig.~\ref{fig:W-DiT}) instead conditions each step
on a deterministic rigid transformation derived from $J$, warping the latent so that only the newly revealed region is generated. Specifically, assuming planar motion on the $XY$-plane (zero vertical velocity and roll/pitch), the inter-frame transformation is $\mathbf{T}^{t+1}_{t} = \exp(\widehat{\xi}_t \Delta t)$,
where $\widehat{\xi}_t \in \mathbb{R}^{4\times 4}$ is the matrix form of the twist induced by the action $a_t = [v_{x,t}, v_{y,t}, \omega_t]^\top$, with planar velocity $\mathbf{v} = [v_{x,t}, v_{y,t}, 0]^\top$ and yaw rate $\boldsymbol{\omega} = [0,0,\omega_t]^\top$.

Applying $\mathbf{T}^{t+1}_{t}$, we forward-warp $\mathbf{z}_t$ to the viewpoint at $t+1$, obtaining $\hat{\mathbf{z}}_{t+1}$. Ego motion inevitably pushes some spatial regions out of the field of view or into occlusion, so we deterministically compute a binary visibility mask $\mathbf{M}_{\text{vis}} \in\{0,1\}^{H\times W}$ from the known geometry alone. However, the inter-frame displacement is small (3--4\,m at $\Delta t = 0.5$\,s), so $\mathbf{M}_{\text{vis}}$ alone leaves the task nearly degenerate: the model can satisfy it by copying the input, which hinders robust representation learning. Following MAE \citep{he2022masked}, we therefore add a random mask $\mathbf{M}_{\text{rand}} \sim \mathcal{B}(1-p)$ with masking ratio $p$, and condition on $\tilde{\mathbf{z}}_{t+1} = (\mathbf{M}_{\text{vis}} \odot \mathbf{M}_{\text{rand}})
\odot \hat{\mathbf{z}}_{t+1}$, where $\odot$ is the Hadamard product and the spatial masks broadcast across $C_z$. Next, instead of fusing the future action into $\tau$ directly
\citep{gu2024dome,shi2025come}, we rasterize it into a spatial signal aligned with index $t+1$. We represent the future actions as physical waypoints $P_t = \{p_{t+i} \in \mathbb{R}^2\}_{i=1}^{\sigma}$ on the BEV plane, anchored in the
ego frame at $t+1$ by cumulatively applying $\mathbf{T}^{t+\sigma}_{t+1}$. Connecting them yields a polyline tracing the intended path onto the future road surface, which
we rasterize to $l_{t+1} \in \{0,1\}^{H\times W\times 1}$ (denoted $I(\cdot)$ in Fig. \ref{fig:W-DiT}). Since warping introduces interpolation error, we concatenate $l_{t+1}$ with $\tilde{\mathbf{z}}_{t+1}$ and refine them with a U-Net before injection, yielding the
conditioning tokens $\hat{f}_{t+1} \in \mathbb{R}^{B\times N\times C_w}$ with $N = H \times W$.

We then inject $\hat{f}_{t+1}$ by token-wise scaling and shifting. Global scale, shift and gate are regressed from the flow timestep $\tau$ as in DiT, while a token-wise projection regresses a spatial control signal from $\hat{f}_{t+1}$ (dual-head injection hereafter); the two are summed per token:
\vspace{-0.2em}
\begin{equation}
[\gamma_{g}, \beta_{g}, \alpha_{g}] = \mathrm{MLP}_{\text{time}}(\tau), \quad
[\Gamma, B, A] = \mathrm{MLP}_{\text{cond}}(\hat{f}_{t+1}), \quad
\gamma^{(i)} = \gamma_{g} + \Gamma^{(i)},
\end{equation}

and analogously for $\beta^{(i)}, \alpha^{(i)}$. On the main stream, $\mathbf{z}^{\tau}_{t+1} = (1-\tau)\epsilon + \tau \mathbf{z}_{t+1}$ is tokenized to $f_{t+1} \in \mathbb{R}^{B\times N\times C_w}$, and the MHSA block becomes
$f'_{t+1} = f_{t+1} + \alpha_1 \odot \mathrm{MHSA}(\gamma_1 \odot \mathrm{LN}(f_{t+1}) + \beta_1)$, where $\gamma_1 = [\gamma^{(1)},\ldots,\gamma^{(N)}]^\top \in
\mathbb{R}^{N\times C_w}$ stacks the token-wise parameters, with $\beta_1, \alpha_1$ defined analogously and the same scheme applied in the subsequent MLP layer. Each W-DiT block therefore modulates every spatial token by its own warped condition, and
the network outputs the instantaneous velocity field $v_\theta(\mathbf{z}^{\tau}_{t+1}, \tau, c)$, where $c$ denotes all conditions.

Following the success of flow models in high-quality generation with few sampling steps \citep{song2025hume,zhang2025epona}, we optimize W-DiT with a velocity regression objective. This objective is sound but not sufficient for long-horizon
rollout: supervising only in the continuous latent space leaves errors that are small in latent norm yet consequential once decoded, which will be re-injected at every autoregressive step. Empirically, models trained with the velocity loss alone match the full objective on the first generated frame but degrade within 10 rollout frames (Tab.~\ref{tab:ablation}).

We therefore introduce an SNR-scaled perception loss. Our insight is to exploit the discrete, semantic nature of 3D occupancy: rather than confining supervision to the continuous
latent manifold, we project the denoised prediction back into occupancy space, redirecting gradient pressure (from CE loss) onto the directions the decoder is sensitive to and anchoring it to a target that does not drift with the latent. The overall objective is
\vspace{-0.2em}
\begin{equation}
 \mathcal{L}_{\mathrm{total}}(\theta) = \mathbb{E}_{\tau, \epsilon, z_{t+1}, \mathbf{c}} \bigg[ \left\| v_\theta - (z_{t+1} - \epsilon) \right\|_2^2 + \lambda \tau^2\frac{\mathbf{M}_{\mathrm{mask}}}{|\mathbf{M}_{\mathrm{mask}}|}\mathrm{CE}\bigg(\mathcal{O}_{t+1}, \mathcal{D}(z_{t+1}^{\tau} + (1-\tau)v_\theta)\bigg) \bigg]
\label{eq:total_loss}
\end{equation}

where $\mathbf{M}_{\text{mask}} = \mathbf{M}_{\text{vis}} \odot
\mathbf{M}_{\text{rand}}$, $\lambda$ balances the two terms, and
$\hat{\mathbf{z}}_{t+1} = \mathbf{z}^{\tau}_{t+1} + (1-\tau)v_\theta$ is the one-step denoised latent recovered from the current velocity field without sampling, decoded by the frozen VAE $\mathcal{D}$ for a mask-region-only cross-entropy against $\mathcal{O}_{t+1}$. Crucially, the $\tau^2$ weighting concentrates this supervision in the low-noise regime, where the one-step reconstruction is close enough to the target for the cross-entropy to carry usable geometric signal. We ablate the usage of perception loss and SNR weight in rows 4 and 5 of Tab. \ref{tab:ablation}.

\subsection{Multi-kilometer road map fusion from single frames}
\label{sec:fusion}

As detailed in Sec. \ref{sec:w-dit}, our W-DiT is capable of stably rolling out thousands of static map frames. While the stochastic nature of flow models effectively bypasses the data scarcity bottleneck of traditional HD maps, the generated outputs remain discrete, ego-centric local scenes. Furthermore, naive fusion strategies, such as temporal averaging, tend to aggressively over-smooth crucial high-frequency details during sequence aggregation, because of the temporal inconsistency in the forecasted frames.

Our strategy consists of two components: key-frame based fusion and branch road detection: We first propose a two-pass keyframe-based fusion heuristic (detailed in Alg.~\ref{alg:keyframe_fusion} in Appendix~\ref{sec:appendix_algorithms}). Given local occupancy predictions and their global poses, we first aggregate a minimally overlapping keyframe subset to establish a sharp foundational map $\mathcal{M}_{\text{global}}$. Subsequently, after a gravity alignment step, the redundant non-keyframes are used to robustly inpaint spatial gaps via threshold-based voting and morphological cleaning.

To enable unbounded map expansion, we systematically identify valid extensible frontiers within $\mathcal{M}_{\text{global}}$. By applying morphological skeletonization to the road manifold, we extract a topological graph to isolate candidate branch points (Alg.~\ref{alg:branch_detection} in Appendix~\ref{sec:appendix_algorithms}). A Dual-Probing Filtering mechanism—comprising topological and 3D semantic probes—is then employed to reject collision-prone directions. By selecting two valid disjoint endpoints and connecting them via a collision-free spline, we prompt W-DiT to autoregressively synthesize the missing intermediate occupancy. This autoregressive bridging paradigm closes the loop, enabling the unbounded, procedural generation of diverse autonomous driving scenarios from a single prompt occupancy.

\begin{figure}[tb]
    \centering
    \includegraphics[width=\linewidth]{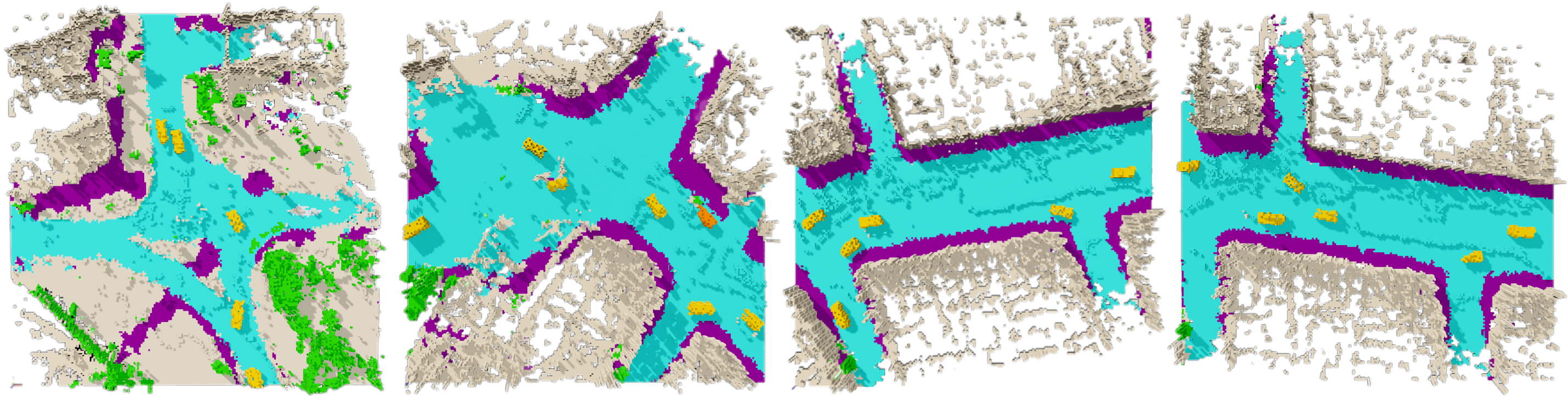}
    \caption{\textbf{Agent initialization within generated static maps.} Each panel displays a 200x200 voxel crop centered on the (unrendered) ego-vehicle from $\mathcal{M}_{\text{global}}$. The first two panels show distinct generated scenes, the third and fourth panels demonstrate the model's multimodality: two different plausible layouts generated for the same revisited location. Orange vehicles (second panel) are stationary.}
    \label{fig:with_agent}
\end{figure}

\subsection{Agent generation and control on top of  static road map}
\label{sec:agent}

Having constructed the city-scale static map $\mathcal{M}_{\text{global}}$ (with explicit lane graphs extracted via Alg. \ref{alg:lane_topology}), we next populate the environment with plausible initial agent states. Unlike log-replay methods, our topologies are procedurally generated, necessitating a generative approach for traffic layouts.

Formally, our objective is to model the conditional probability distribution of the agent layout given 1 frame of static environment: $P(\mathcal{A} | \mathbf{z}_{t})$, where $\mathcal{A} = \{(x_i, y_i, s_i)\}_{i=1}^N$ denotes the set of vehicle centers $(x_i, y_i)$ motion states $s_i \in \{\text{dynamic}, \text{static}\}$ and $\mathbf{z}_{\text{static}}$ is cropped from a previously fused large map given a random pose. To make this distribution tractable for our flow-based generative framework, we map the discrete set $\mathcal{A}$ into a continuous 2D spatial heatmap $\mathcal{H}$, encoding static and dynamic vehicles as positive and negative standard Gaussian kernels, respectively. Then we train a compact Diffusion Transformer (DiT-S) to learn the continuous distribution $P(\mathcal{H} | \mathbf{z}_{t})$. To prevent the DiT-S from overfitting and memorizing small-scale dataset layouts, we apply multiple spatial augmentations during training (detailed in Appendix)

During inference, the sampled heatmap $\mathcal{H}$ is discretized via Non-Maximum Suppression (NMS) to extract individual vehicle instances. Each vehicle is then projected onto the nearest lane to establish a kinematically feasible initial yaw. In figure \ref{fig:with_agent}, we visualize several samples after add the agents to their initial pose. Finally, we employ A* routing and the extended 2D Intelligent Driver Model (2D-IDM) \cite{kesting2010enhanced} to roll out microscopic kinematic trajectories toward randomly selected junctions in $\mathcal{M}_{\text{global}}$ (Alg.~\ref{alg:idm_rollout} in Appendix~\ref{sec:appendix_algorithms}). While we utilize classic heuristics to validate simulation feasibility, our  initialized map topology and agent poses guarantee plug-and-play compatibility with any modern learning-based trajectory forecasting model.

\section{Experiment}

We organized our experiments to answer the following 3 questions: Q1. \textbf{Static Realism \& Stability}: Can the proposed W-DiT structure generate highly realistic and spatiotemporally consistent static environments, especially over extended horizons? Q2. \textbf{Generation Diversity}: Does OccSim exhibit sufficient multimodality and diversity in long-horizon rollout? Q3. \textbf{Dynamic Fidelity \& Downstream Utility}: Can OccSim be used to train temporal-based tasks like 4D semantic occupancy forecasting more effectively than data collected from CARLA?

\subsection{Metrics Protocol Design}
\label{sec:metrics}

No single metric suffices to evaluate log-free generation: beyond the logged horizon no paired ground truth exists, so realism can only be measured against the marginal distribution, which a model repeatedly emitting one plausible scene would satisfy , while diversity scores are themselves inflated by structural breakdown. We therefore report fidelity, diversity and road-network connectivity jointly, reference them against an explicit collapse baseline, and ground the protocol in an external downstream task (\textbf{Q3}) that the generator cannot optimise against.

Specifically, to answer \textbf{Q1}, we assess its performance across two primary dimensions: the realism of the generated static environments and the dynamic fidelity of the traffic simulation. Following previous works \cite{bian2024dynamiccity, yang2025x}, we evaluate static realism via 2D/3D-FID, KID, and MMD (denoted by $\textrm{Dis}$ below).

To formalize realism, we define two evaluation settings to measure the performance of model under different context: \textbf{1. Conditional fidelity}: ${\textrm{Dis}\left( P_{\theta}(\hat{\mathcal{O}}_t | \mathcal{O}_0, J), P_{data}(\mathcal{O}_t | \mathcal{O}_0, J) \right)}$. This evaluates the structural accuracy at each specific time index $t \in \{1, \dots, \mathcal{K}\}$ ($\mathcal{K}$ equal to the length of GT sequence) under the strict constraint of the initial observation $\mathcal{O}_0$ and ego-trajectory $J$. \textbf{2. Unconditional realism}: $\textrm{Dis}\left( P_{\theta}(\hat{\mathcal{O}}_t | \mathcal{O}_0, J), P_{data}(\mathcal{O}) \right)$, where t depend on the rollout length $K$. Although the generation inherently relies on the initial observation $\mathcal{O}_0$ and trajectory $\mathbf{J}$, long-horizon rollouts can reasonably diverge into various valid structures. Therefore, instead of comparing against the strictly paired future frame, we measure the distance between the conditionally generated samples at time $t$ and the global marginal distribution of the entire ground truth training set $\mathcal{O}$. This ensures long-term rollouts universally reside on the valid data manifold.

In \textbf{Q2}, we introduce the Vendi score \cite{friedman2022vendi} and Semantic IoU diverse score. Specifically, the Semantic IoU diverse score evaluates structural divergence in the decoded semantic space by calculating the complement of the average Intersection over Union across multiple stochastic rollouts generated from the same initial scene. Meanwhile, the Vendi score operates in the latent feature space—extracted via a pretrained encoder—to quantify the effective sample size of the generated distributions. These metrics reflect the model's multimodality, where higher scores indicate a richer variety of plausible future layouts, demonstrating the simulator's capability to model environmental uncertainty.

To address \textbf{Q3}, we evaluate the dynamic simulation performance through a zero-shot transfer setting. To the best of our knowledge, we are the first to demonstrate the downstream utility of a data-driven simulator by evaluating its pre-training efficacy on the 4D semantic occupancy forecasting task. Specifically, we utilize continuous data streams collected directly from the ego-vehicle's perspective within our generated simulation. Performance on this downstream task serves as a  proxy, directly reflecting the simulator's capability to faithfully emulate complex, real-world traffic dynamics.

Regarding datasets, we utilize Occ3D-nuScenes \cite{tian2023occ3d} for all static generation realism comparisons to ensure fairness in training data volume. For the remaining experiments, we employ UniOcc-Waymo \cite{wang2025uniocc} or UniOcc-nuScenes due to their unified voxel categorization. Additional simulation metrics and training/implementation details are provided in Appendix~\ref{sec:appendix_training_details}.

\subsection{Main results comparisons}

\begin{figure}[t]
    \centering
    \includegraphics[width=\linewidth]{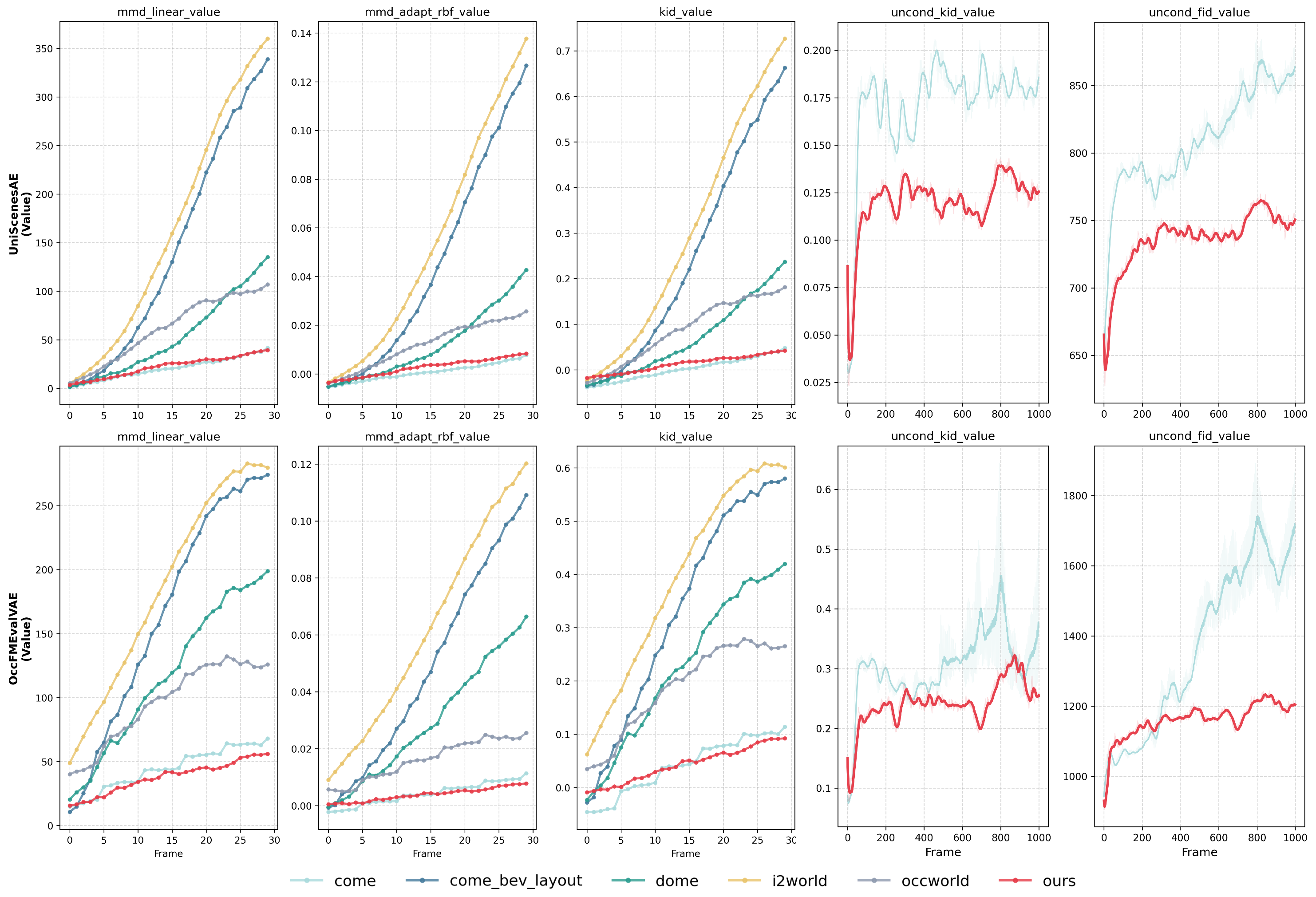}
    \caption{Realism evaluation of generated 3D static occupancy. The left three columns report the conditional fidelity (MMD and KID) under a 30-frame constraint. The right two columns showcase the unconditional realism (KID and FID) over a 1000-frame long-horizon rollout. Our W-DiT-based method obtains the best long-horizon stability and conditional realism.}
    \label{fig:cond_3d}
\end{figure}

We first evaluate the realism and stability of the generated static road maps. In Fig. \ref{fig:cond_3d}, we compare our results with previous occupancy world models in terms of both conditional fidelity and unconditional realism. For a comprehensive analysis, our evaluations are conducted across two distinct VAE latent spaces (derived from UniScene \cite{li2025uniscene} and OccFM \cite{liu2025towards}). As illustrated in the figure, our method (shown in red) demonstrates superior long-term stability and structural realism, significantly outperforming baseline models that suffer from rapid error accumulation. Furthermore, due to the limited number of ground truth (GT) samples (fewer than 1000 per timestep $t$), we specifically selected unbiased metrics—namely KID and MMD with two different kernels—to assess conditional fidelity. Here, 1000 frames rollout generated conditioned on straight trajectory after GT's, with constant velocity. 
 
\begin{figure}[t]
    \centering
    \includegraphics[width=\linewidth]{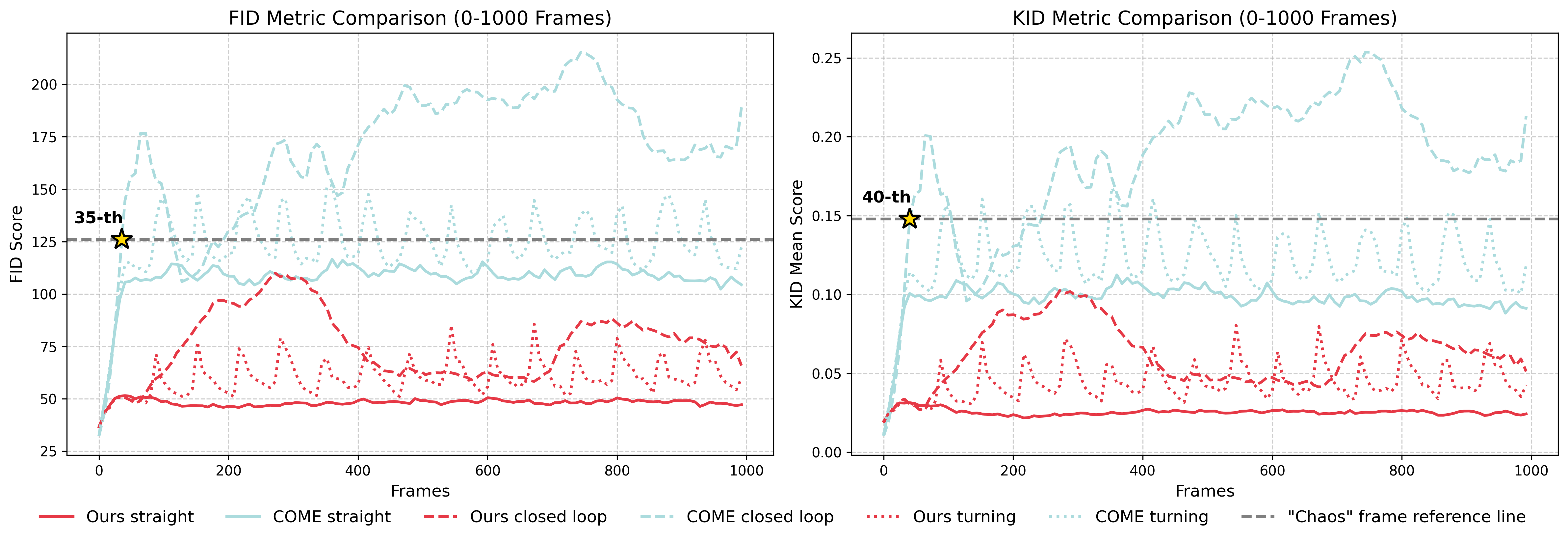}
    \caption{Long-horizon 2D realism under varying trajectories. FID and KID comparisons for 2D projections of the 3D generated occupancy. We evaluate three ego-actions: straight, closed-loop, and continuous turning. The dashed reference line indicates the ``chaos'' threshold characterizing scene collapse. In all scenarios, our model (red) remains strictly below this threshold, vastly outperforming the previous SOTA model(which exceed the ``chaos'' threshold in 35/40-th frame generated) and ensuring long-term structural integrity regardless of the driving action. }
    \label{fig:uncond_2d}
\end{figure}

This performance gap observed in the 3D occupancy metrics becomes even more pronounced when evaluating realism on the 2D plane. As illustrated in Fig. \ref{fig:uncond_2d}, to increase the rigor of our evaluation, we selected three distinct trajectories to compare our method against the previous state-of-the-art diffusion based occupancy world model, COME (details regarding the specific trajectory shapes are provided in Fig.~\ref{fig:traj_vis} of the Appendix). Notably, even when evaluated on the most challenging curved and closed-loop trajectories, our approach still outperforms the previous state-of-the-art method's results on simpler straight-line trajectories.

Furthermore, to establish a lower-bound reference, we constructed a 1000-frame occupancy map collection to serve as a ``chaos'' baseline. This was achieved by randomly selecting structurally collapsed samples from the model's output that lack coherent semantic information. Qualitatively, any metric score surpassing this baseline threshold indicates a completely unusable state. Visualizations of this chaotic baseline collection are available in Appendix~\ref{sec:appendix_chaos}. 

\begin{figure}[h]
    \centering
    \includegraphics[width=\linewidth]{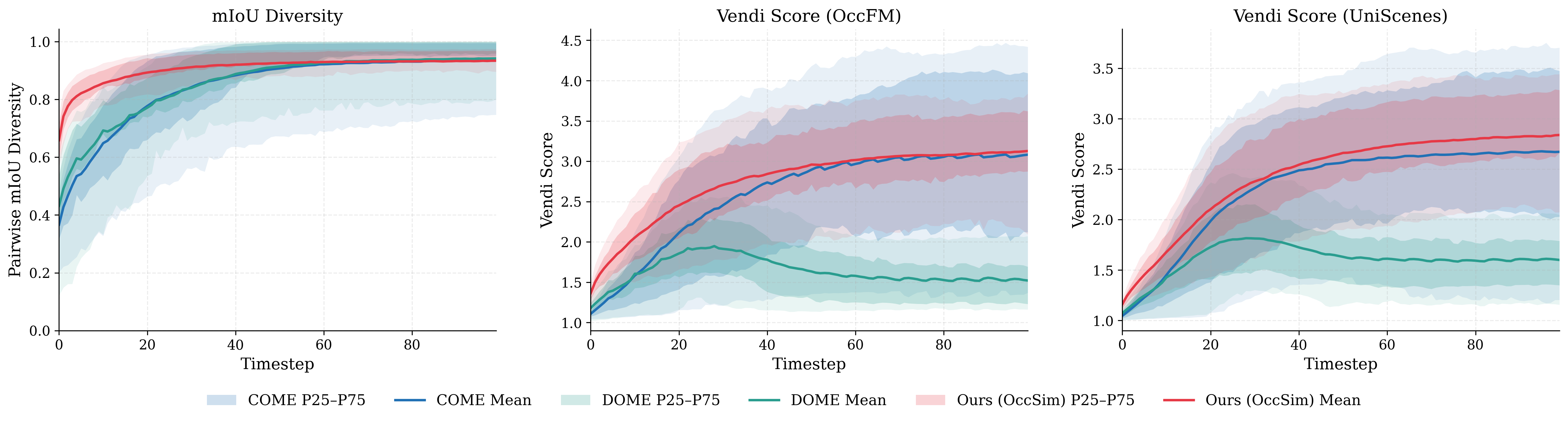}
    \caption{Generation diversity comparison with previous occupancy world model. Pairwise mIoU diversity and Vendi scores evaluated over 100 timesteps using \textbf{10 random seeds}. Solid lines and shaded areas denote the mean and the 25th-75th percentiles, respectively. Our methods performed the best diversity in the rollout among all tested generative method.}
    \label{fig:diversity}
\end{figure}

To evaluate the generation diversity over long-term rollouts, we analyze the Pairwise mIoU Diversity and Vendi Scores across first 100 timesteps, as illustrated in Fig. \ref{fig:diversity}. Our proposed method consistently outperforms prior generative baselines. Notably, at early timesteps, our approach demonstrates a rapid increase in diversity, establishing a significant margin over both COME and DOME. As the autoregressive generation extends, our method maintains a high and stable level of diversity without performance degradation. In both OccFM and UniScenes latent spaces, our method achieves the highest Vendi Scores, indicating a richer variety of generated geometric and semantic structures. 

\begin{table}[h]
\centering
\caption{Diversity and fidelity comparison with other non-world-model occupancy generation methods across distances. SemCity and InfiniCubes extend map boundaries via unconditional / HD-Map-conditioned diffusion outpainting, where quality degrades with extrapolation distance. UniScene$^{\dagger}$ requires a aligned per-frame HD-Map to generate occupancy and is thus distance-independent,  we treat it as a performance upper bound; results of condition-free version also reported below. $\star$: SemCity's released code does not support Occ3D, we reproduced results with public released code base. LCC: Largest connected components, measures the road-network completeness of the stitched map.  All experiments are run under \textbf{5 different seeds} and 150 scenes then averaged. Details and visualization can be found in Appendix \ref{sec:train_details} and \ref{sec:comp_outpaint}.}
\label{tab:physical_distance}
\resizebox{\textwidth}{!}{
\begin{tabular}{llcccccccc}
\toprule
\multirow{2}{*}{Model} & \multirow{2}{*}{Dataset} & \multirow{2}{*}{HD-Map?} & \multicolumn{6}{c}{2D-FID $\downarrow$ / Mean Vendi Score at different physical distance $\uparrow$} & \multirow{2}{*}{\makecell{LCC. Ratio $\uparrow$}} \\
\cmidrule(lr){4-9}
 & & & 40m & 400m & 1km & 2km & 5km & 10km & \\
\midrule
SemCity\cite{lee2024semcity}$^{\star}$ & \multirow{3}{*}{\makecell{Occ3d-\\NuScenes}} & \ding{55} & 46.31/1.00 & 97.07/2.03 & 102.01/2.04 & 147.36/2.06 & 141.96/1.93 & 148.29/1.95 & $11.7 \%$ \\
\cellcolor{gray!25}UniScene\cite{li2025uniscene}$^{\dagger}$ & & \cellcolor{gray!25}\ding{51} & \multicolumn{6}{c}{\cellcolor{gray!25} 33.12/1.55} & \cellcolor{gray!25}-- \\
\cellcolor{gray!25}UniScene\cite{li2025uniscene} & & \cellcolor{gray!25}\ding{55} & \multicolumn{6}{c}{\cellcolor{gray!25} 215.59/3.82} & \cellcolor{gray!25}-- \\

Ours & & \ding{55} & 36.48/1.24 & 50.00/2.05 & 50.89/2.22 & 46.40/2.27  & 48.60/2.29 & 47.13/2.35 & $92.9\%$ \\
\bottomrule
\end{tabular}
}
\label{tab:outpainting}
\end{table}

In Tab. \ref{tab:physical_distance}, we further compare our world model based static occupancy generation method with previous out-painting based solution. Although temporal rollout introduces error accumulation that spatial outpainting does not face, W-DiT still surpasses all outpainting-based baselines in terms of both fidelity (FID) and diversity (Vendi Score), and the gap widens with distance. It is worth noting that both of the 10km spatial distance in Tab. \ref{tab:physical_distance} (about 3000 frames) in or 1000 temporal frames in Fig. \ref{fig:uncond_2d} are the upper bound we can quantitatively assess given the compute cost. However, we find that W-DiT consistently maintains stable generation quality on a scale of more than two orders of magnitude. In Sec. \ref{sec:100km}, we qualitatively present the \textbf{100km+ city scale occupancy generated from OccSim.} As far as we
know, this is the largest scale tested in occupancy based driving scenario generation.

To validate the practical utility of our generated world: we then use 4D semantic forecasting as a proxy task to show the downstream utility of using OccSim as data generator. For this experiment, we configure the model to unroll along a curved, closed-loop trajectory, expanding the generation horizon from 1,000 to 3,000 frames to construct a significantly larger map. Based on this extended map, we build the simulation environment utilizing the methodologies detailed in Sec. 3.2 and Sec. 3.3. As demonstrated in Tab. \ref{tab:results_trans}, even when relying on a relatively straightforward dynamics algorithm such as the 2D IDM to control agents, OccSim provides better training signals for simulating overall traffic flow and background dynamics than CARLA. Notably, models trained on data generated by OccSim achieve better performance compared to those trained on the CARLA simulator (Uni-C), showing an improvement of at least $14\%$ relative in average mIoU across the two semantic forecasting models. When we collect 5x more data from UniOcc to train the downstream model, this gap further expands to $52\%$. In terms of 3s average IoU, our method reaches 67\% and 72\% of the upper-bound's performance. Improvement can also be observed from the alignment of the intent of the human driver measured by the L2 distance. The same ordering holds for the open-loop planning L2 of OccWorld's ego trajectory head, all using a non-learned IDM controller. Augmenting the full real Waymo logs with an extra 30\% of OccSim data improves the average 1--3s IoU by a further 7\% relative (Sec.~\ref{sec:augmentation}), showing OccSim also complements real data.

\begin{table}[tb]
  \centering
  \caption{The performance of different semantic forecasting model trained with occupancy collected from UniOcc-Carla(Uni-C) or OccSim, test on UniOcc-NuScenes(Uni-N) validation set. mIOU measured over all background occupancy categories and car category for foreground. $\dagger$: We collected the same amount of data on OccSim as that provided by UniOcc for Carla for fairness. $\star$: Use $\times5$ amount of data collected from OccSim to train downstream models. Stoch.: whether the model is stochastic. The gray-shaded rows represent the performance upper bound, as the models are trained and tested on the same domain without domain shift. We omit the L2 error for OccFM here as it's future action conditioned. }
  \label{tab:results_trans}

  \begin{tabularx}{\textwidth}{l lYYYYYYYYYY Y}
    \toprule
    \multirow{2}{*}{Model} & \multirow{2}{*}{Train Data} & \multicolumn{3}{c}{mIoU} & \multicolumn{3}{c}{IoU} & \multicolumn{3}{c}{L2 (m)} & \multirow{2}{*}{$\text{IoU}_{\text{veh}}$} & \multirow{2}{*}{Stoch.} \\
    \cmidrule(lr){3-5} \cmidrule(lr){6-8} \cmidrule(lr){9-11}
    &  & 1s & 2s & 3s & 1s & 2s & 3s & 1s & 2s & 3s &  & \\
    \midrule
    \rowcolor{gray!15} OccWorld\cite{zheng2024occworld} & Uni-N & 26.53 & 16.73 & 12.33 & 32.17 & 22.88 & 18.41 & 0.43 & 1.08 & 1.99 & 17.10 & \ding{55} \\
    OccWorld & Uni-C & 11.79 & 8.35 & 6.75 & 18.31 & 14.09 & 12.03 & 0.72 & 2.31 & 4.05 & 6.58 & \ding{55} \\
    $\text{OccWorld}^{\dagger}$ & OccSim & 13.51 & 9.56 & 7.61 & 20.53 & 15.48 & 13.24 & 0.65 & 1.42 & 3.27 & 7.46 & \ding{55} \\
    $\text{OccWorld}^{\star}$ & OccSim & 19.54 & 12.31 & 9.08 & 23.87 & 16.92 & 13.47 & 0.61 & 1.38 & 2.91 & 12.41 & \ding{55} \\
    \midrule
    \rowcolor{gray!15} OccFM\cite{liu2025towards} & Uni-N & 36.28 & 25.10 & 19.47 & 41.22 & 30.90 & 25.02 & - & - & - & 28.79 & \ding{51} \\
    OccFM & Uni-C & 6.99 & 5.34 & 4.70 & 8.63 & 6.50 & 5.71 & - & - & - & 6.42 & \ding{51} \\
    $\text{OccFM}^{\dagger}$ & OccSim & 15.93 & 11.76 & 9.26 & 18.31 & 14.10 & 11.75 & - & - & - & 13.28 & \ding{51} \\
    $\text{OccFM}^{\star}$ & OccSim & 25.32 & 18.96 & 13.10 & 29.19 & 21.08 & 13.93 & - & - & - & 15.82 & \ding{51} \\
    \bottomrule
  \end{tabularx}
\end{table}

\subsection{Ablation study}
\label{sec:ablation}

In Tab. \ref{tab:ablation}, we shows the three components are only effective jointly. Token-wise condition injection alone (row~3) does not stabilise the rollout and is in fact worse than the plain baseline at intermediate horizons; adding the perception loss on top of it (row~4) improves the 500-frame endpoint but still leaves quality degrading monotonically. Only when the perception loss is SNR-scaled (row~5) does the rollout become stable, halving the 500-frame FID relative to row~3 and keeping it within $12\%$ of the first generated frame. This matches the role of the $\tau^2$ weighting: it concentrates occupancy-space supervision in the low-noise regime, where the decoded prediction is close enough to the GT for the cross-entropy term to carry geometric signal.

To evaluate agent motion realism under our log-free setting, following Scenario Dreamer~\citep{rowe2025scenario}, Tab.~\ref{tab:agent_motion_jsd} reports the JSD between simulated and real agent motion, with a GT-vs-GT as an empirical lower bound. We decompose the remaining gap in three steps. Replacing real trajectories with IDM control on lane graphs extracted from GT occupancy (row 2) yields the largest increase. This step bundles two costs: the simplification of the IDM itself, and the representational cost of recovering topology from voxelized semantics rather than from a vectorized map; we analyze the latter in Sec.~\ref{sec:heuristic}, where lane graphs extracted from GT semantic occupancy are compared against those rasterized from HD maps and the residual Chamfer discrepancy remains sub-voxel relative to the $0.4$m grid. Substituting GT agent positions with our Layout Generator (row 3) adds a smaller increment, so initialization is a secondary error source. Most importantly, moving from real to generated occupancy (row 4) changes the JSD only marginally across all four behaviour statistics, and is in fact lower on nearest-distance: our map topology is therefore not the limiting factor, and the remaining gap is dominated by the representation-and-controller stage, i.e.\ exactly the component our pipeline exposes as replaceable.

Tab. \ref{tab:safety_eval} provide a collision rate v.s. simulation time analyze, and show that initialization from our Layout generator provide a more stable long horizon simulation results. Also in Tab. \ref{tab:speed_breakdown}, we measure the time taken by OccSim components. Given a 1000-step action and initial frame, building a scene need encode the first frame, runs 1000 steps of W-DiT rollout, decodes the results, and heuristic to extract lanes (around 656s). During simulation, if the ego-vehicle is still in pre-computed area, the simulation latency is equal to IDM planning time (7.1ms), while after every 40m, frontal-view layout recomputation is required, which encodes the front area, generates a layout, and performs IDM-based planning (92.68ms).

Furthermore, we report two additional ablations in the appendix. In Sec.~\ref{sec:nfe}, we ablate the number of function evaluations (NFE) jointly with the single-step rollout stride. In Sec.~\ref{sec:sphere}, we analyze the semantic readout of our VAE in detail and show that it is crucial for long-horizon stability; we attribute this to the implicit constraint that the spherical readout imposes on the latent manifold during training. As this design is orthogonal to the world model itself, it can be adopted by any occupancy world model to obtain an $11\%$ relative improvement at negligible cost.

\begin{table}[t]
    \centering
    \small
    \setlength{\tabcolsep}{5pt}
    \caption{W-DiT design ablation, reported as the UniScene-AE 3D unconditional realism FID score at the $n$-th generated frame. All three proposed components are essential for long-horizon stability. When not using dual-head injection, we adopt the classical spatiotemporal architecture in DOME.}
    \label{tab:ablation}
    \resizebox{\linewidth}{!}{%
    \begin{tabular}{c|ccc|ccccccc}
        \toprule
        Architecture & \makecell{Percep.\\Loss} & \makecell{Dual-head\\injection} & \makecell{SNR\\scaled} & 1st & 5th & 10th & 30th & 50th & 100th & 500th \\
        \midrule
        \multirow{2}{*}{Spatiotemporal DiT} & \ding{55} & \ding{55} & \ding{55} & 663.89 & 712.56 & 789.91 & 1023.51 & 1281.99 & 1587.21 & 1612.40 \\
        & \ding{51} & \ding{55} & \ding{51} & 661.79 & 710.63 & 801.11 & 1034.75 & 1342.72 & 1601.15 & 1607.97\\

        \midrule
        \multirow{3}{*}{W-DiT} & \ding{55} & \ding{51} & \ding{55} & 672.15 & 756.28 & 1013.12 & 1335.88 & 1295.42 & 1378.56 & 1526.95 \\
         & \ding{51} & \ding{51} & \ding{55} & 696.13 & 833.17 & 946.23 & 1306.21 & 1415.59 & 1427.66 & 1440.57 \\
         & \ding{51} & \ding{51} & \ding{51} & 665.23 & 630.34 & 645.96 & 687.90 & 698.87 & 712.65 & 744.71 \\
        \bottomrule
    \end{tabular}%
    }
\end{table}

\begin{table*}[!ht]
\centering
\begin{minipage}[t]{0.48\textwidth}
\centering
\setlength{\tabcolsep}{4pt}
\renewcommand{\arraystretch}{1.15}
\caption{Agent motion realism under our log-free setting. We adopt the behaviour-simulation JSD protocol of Scenario Dreamer~\citep{rowe2025scenario} (Lin.\ Spd., Ang.\ Spd., Acc., Near.\ Dist.). Most of the discrepancy comes from IDM control and layout initialization rather than the generated occupancy maps. \textbf{GT vs.\ GT split} (real trajectories randomly split in half) serves as an empirical lower bound; \textbf{IDM-on-real-map} isolates the effect of our generated road topology by running the \emph{same} IDM controller on real lane graphs from GT occupancy. Behav. and Occ. : Agent action behavior and occupancy used to extract lanes for IDM control.}
\label{tab:agent_motion_jsd}
\footnotesize
\resizebox{\linewidth}{!}{%
\begin{tabular}{@{}lllcccc@{}}
\toprule
\multirow{2}{*}[-3.5pt]{Layout} & \multirow{2}{*}[-3.5pt]{Behav.} & \multirow{2}{*}[-3.5pt]{Occ.}
& \multicolumn{4}{c}{Distributional JSD ($\times 10^{-2}$) $\downarrow$} \\
\cmidrule(lr){4-7}
 & & & \makecell{Lin.\\Spd.} & \makecell{Ang.\\Spd.} & Acc. & \makecell{Near.\\Dist.} \\
\midrule
\rowcolor{gray!15}
\makecell[l]{GT split\\(l.b.)} & GT & GT
& 1.42 & 0.05 & 7.61 & 0.15 \\
\midrule
GT & IDM & GT
& 5.91 & 0.13 & 11.76 & 0.7 \\
\makecell[l]{Layout\\Gen.} & IDM & GT
& 8.13 & 0.35 & 17.9 & 1.7 \\
\rowcolor{blue!8}
\makecell[l]{Layout\\Gen. (Ours)} & IDM & Gen.
& 8.76 & 0.35 & 18.6 & 1.4 \\
\bottomrule
\end{tabular}%
}
\end{minipage}
\hfill
\begin{minipage}[t]{0.48\textwidth}
\centering
\caption{Closed-loop Interaction Safety Evaluation (Collision Rate \%). Each setting tested using 7500 samples in 150 maps. We use max 6 agents to collect 10s non-collision video clips for downstream training.}
\label{tab:safety_eval}
\footnotesize
\resizebox{\linewidth}{!}{%
\begin{tabular}{@{}lcccccc@{}}
\toprule
\textbf{Init. method} & \textbf{Max agents}
& \textbf{3s} & \textbf{5s} & \textbf{10s}
& \textbf{20s} & \textbf{60s} \\
\midrule
Ours
 & 4 & 0.00 & 0.00 & 0.12 & 0.45 & 0.49 \\
 & 5 & 0.00 & 0.03 & 0.23 & 0.42 & 0.61 \\
 & 6 & 0.02 & 0.11 & 0.38 & 0.44 & 0.77 \\
 & 7 & 0.05 & 0.11 & 0.41 & 0.52 & 0.79 \\
Random
 & 7 & 0.19 & 0.23 & 0.71 & 0.91 & 1.00 \\
\bottomrule
\end{tabular}%
}
\vspace{2em}
\renewcommand{\arraystretch}{1.15}
\caption{Breakdown of inference speed per frame (ms). All results are tested on 1$\times$ RTX4090 with float32 precision. No further acceleration like pre-compile applied here.}
\label{tab:speed_breakdown}
\footnotesize
\resizebox{\linewidth}{!}{%
\begin{tabular}{@{}lccccc@{}}
\toprule
& \textbf{Encoder} & \textbf{W-DiT} & \textbf{Decoder} & \makecell{\textbf{Layout}\\\textbf{Gen.}} & \makecell{\textbf{IDM Plan.}\\\textbf{\& rendering}} \\ \midrule
\textbf{Latency} (ms) & 16.88 & 386.4 & 9.49 & 68.7 & 7.1 \\ \bottomrule
\end{tabular}%
}
\end{minipage}
\end{table*}

\section{Conclusion}
In this paper, we present OccSim, the first occupancy world model-driven simulator capable of generating diverse and realistic rollouts spanning thousands of frames without any BEV layout or HD map as input. We demonstrate that a comprehensive simulator built upon such generated road topologies can replicate the spatiotemporal dynamics of driving scenarios with significantly higher fidelity than traditional simulators. As our framework is inherently compatible with complex learning-based policies, future work could study replacing IDM with other data-driven control strategies, alongside extending the simulator to support high-fidelity multi-modal generation.

\subsection*{Acknowledgments}
This research was enabled in part by the Digital Research Alliance of Canada (\texttt{alliancecan.ca}), the NVIDIA Academic Grant Program, and Google TPU Research Cloud (TRC).

\clearpage\clearpage
\section*{Appendix}
In this appendix, we supplement the following materials to support the findings and conclusions drawn in the main body of this paper.

\startcontents[appendices]
\printcontents[appendices]{l}{1}{\setcounter{tocdepth}{3}}

\newpage

\section{Training and metric details}
\label{sec:training_and_metrics_details}

\subsection{Details of OccSim configuration}
\label{sec:train_details}

For training of OccSim, Occ3D-nuScenes \citep{tian2023occ3d} is used for all static realism comparisons to equalise training data volume; Considering the training cost, for performance reported in Fig. \ref{fig:cond_3d}, we use the public released weight trained on original Occ3d NuSc dataset (with dynamic agent) for all conditional evaluation (First 3 figures on the top), where we can see a close performance between W-DiT and COME. To fairly compare W-DiT with the previous SoTA occupancy world model, we used the preprocessed static occupancy retrained COME in all unconditional experiments in Fig. \ref{fig:cond_3d}.

For Tab.~\ref{tab:results_trans}, OccSim is trained on UniOcc-Waymo~\citep{wang2025uniocc}, and downstream forecasting models trained on the data collected from it are evaluated on the UniOcc-nuScenes validation set. We adopt UniOcc labels rather than Occ3D for their unified voxel categorization; since no Occ3D-nuScenes label is used to train OccSim, no label leakage occurs at test time. The OccSim-collected data ($\dagger$) matches UniOcc-Carla in volume, and $5\times$ ($\star$) denotes five times this volume collected from OccSim, with OccSim's own training data unchanged.

The two components OccSim's: W-DiT and the Layout Generator, was conducted separately as follows. For W-DiT, we trained for 200 epochs on 4 Blackwell RTX Pro 6000 GPUs with a batch size of 32. Training on the Nuscenes dataset took approximately 100 GPU hours. For the Layout Generator, we trained for 500 epochs on 8 A100 80G GPUs, taking approximately 160 GPU hours. All experiments used AdamW with cosine annealing, which decayed the learning rate from 3.2e-5 to 3.2e-6.

In Eq.~\ref{eq:total_loss}, we set $\lambda=2$, with 20\% probability to drop condition during training for classifier-free guidance \citep{ho2022classifier} in sampling stage. When generating $\mathbf{M}_{\text{rand}}$, we control the maximum of overall masked region $\mathbf{M}_{\text{mask}}$ between 10\% to 50\% of the full latent size. $\tau$ in Eq.~\ref{eq:total_loss} is sampled from $\text{sigmoid}(\mathcal{N}(0, I))$. In all W-DiT-based rollout experiments described in the main text, we used NFE=10, CFG=1.1, and a roll-out of 3m per step. We further analyze these hyperparameters in Sec.~\ref{sec:nfe}.

\subsection{Training details of baseline reproduced}

For SemCity~\citep{lee2024semcity} reproduced, we first retrain the triplane autoencoder in existing codes on preprocessed Occ3d-Nusc datasets (filter all of dynamic agents) for 30 epochs, the rIOU/rMIOU is 86.07\%/76.7\% for Occ3d-Nusc. Then the triplane diffusion was trained with these triplane representations for 50000 steps and 70000 steps for Occ3d-Nusc and Occ3d-Waymo, with batch size 18, initial learning rate $5\times10^{-4}$ linearly annealed to 0, EMA rate 0.9999. Final training losses were 0.126–0.129 (vs. $\sim0.78$ at initialization) for both datasets. All other architectural hyperparameters (embedding dim, MLP depth, triplane channel width, DDPM schedule) match the original SemCity configuration. 


\subsection{Augmentation in Layout generator}
\label{sec:appendix_augmentations}

Theoretically, the distribution of valid vehicle behaviors given a specific map representation, denoted $P(\mathcal{H} | \mathbf{z}_{t})$, is inherently highly multimodal, as various driving intentions can perfectly align with the same static environment. However, direct optimization on a finite training dataset $\mathcal{D} = \{(\mathbf{z}_{t}^{(i)}, \mathcal{H}^{(i)})\}_{i=1}^{N}$ poses a severe risk of overfitting, as the number of samples is only 28K to train the DiT-S. In practice, for any specific scene $\mathbf{z}_{t}^{(i)}$ in the log, we only observe a single deterministic realization of the vehicles' behavior $\mathcal{H}^{(i)}$. Consequently, without proper regularization, model tend to memorize these exact map-action pairs, effectively collapsing the learned conditional distribution into a Dirac delta function, $P_{\theta}(\mathcal{H} | \mathbf{z}_{t}^{(i)}) \approx \delta(\mathcal{H} - \mathcal{H}^{(i)})$. 

To prevent the model from memorizing spurious correlations between specific voxel arrangements and deterministic position, we introduce the data augmentation in layout generator training. By applying paired spatial transformations $\mathcal{T}$ to both the condition and the target, i.e., $(\mathcal{T}(\mathbf{z}_{t}), \mathcal{T}(\mathcal{H}))$, we artificially expand the support of the empirical distribution. Specifically, we introduce a two-stage data augmentation pipeline comprising agent-level perturbations and global spatial transformations. 

First, to manage scene density and prevent the model from overfitting to overly crowded environments, we cap the maximum number of vehicles per frame at 10; if a frame contains $N > 10$ vehicles, we uniformly sample a subset of exactly $10$ vehicles without replacement, ensuring each vehicle has an equal retention probability of $10/N$. For each retained vehicle, we project it onto a 2D spatial heatmap $\mathcal{H} \in \mathbb{R}^{200 \times 200 \times 1}$ as a Gaussian distribution. Specifically, the Gaussian is centered at the vehicle's spatial coordinates with a kernel radius of 3. To simulate positional uncertainty and enrich the local behavioral manifold, we independently perturb the center of each Gaussian, randomly shifting it within a local $5 \times 5$ neighborhood restricted to the drivable surface area. 

Following this localized perturbation, we apply rigid global transformations to the entire augmented scene (including both the occupancy map and the constructed heatmap $\mathcal{H}$). We uniformly sample and apply discrete rotations $\mathcal{R} \in \{90^\circ, 180^\circ, 270^\circ\}$ and axis-aligned reflections (flipping along the x- or y-axis). Crucially, identical spatial transformations are applied to the corresponding latent representations to maintain spatial alignment with $\mathcal{H}$. While such reflections fundamentally invert the handedness of the traffic rules (effectively synthesizing valid left-hand traffic scenarios from right-hand data), this joint transformation ensures that all local topological constraints and relative vehicle dynamics remain physically coherent. This composite strategy effectively artificially expands the empirical distribution, forcing the network to learn robust, viewpoint-invariant geometric representations.

\subsection{Additional metrics details}
\label{sec:appendix_training_details}

To evaluate the diversity of the generated stochastic rollouts, we employ two distinct metrics: Pairwise mIoU Diversity in the decoded semantic space, and the Vendi Score \citep{friedman2022vendi} in the latent feature space. 

For the decoded semantic space, we generate $N$ independent sequences over $\mathbb{T}$ frames using different random seeds. At any given timestep $t$, we compute the mean Intersection over Union ($\text{mIoU}$) across all semantic categories for every unique pair of generated occupancy grids. The Pairwise mIoU Diversity at time $t$ is then defined as the complement of the average pairwise similarity:
$$D_t = 1 - \frac{1}{|\mathcal{P}|}\sum_{(i,j)\in\mathcal{P}} \text{mIoU}_{t}^{(i,j)}$$
where $\mathcal{P} = \{(i,j) : 1 \le i < j \le N\}$ denotes the set of all unordered seed pairs. The rollout-level diversity is obtained by averaging $D_t$ over time.

To capture structural diversity in the latent space, we report the Vendi Score, which measures the effective sample size of the generated distributions. First, we extract frame-level features $\{x_i\}_{i=1}^N$ using pretrained encoders (i.e., the OccFM VAE encoder and the UniScenes AE encoder). We then compute a cosine similarity kernel matrix $\mathcal{Q}$, where its elements are given by $\mathcal{Q}_{ij} = (1 + \tilde{x}_i^\top \tilde{x}_j)/2$, with $\tilde{x}$ being the $\ell_2$-normalized feature. The Vendi Score is calculated as the exponential of the von Neumann entropy of the normalized kernel matrix $\hat{\mathcal{Q}} = \mathcal{Q}/\text{tr}(\mathcal{Q})$:
$$V = \exp \left( -\sum_\ell \lambda_\ell \log \lambda_\ell \right)$$
where $\{\lambda_\ell\}$ are the non-negative eigenvalues of $\hat{\mathcal{Q}}$.

Intuitively, the Vendi Score quantifies the effective number of distinct modes within a given set of generations. By evaluating the entropy of the similarity eigenspectrum, it properly penalizes highly correlated samples. A higher Vendi Score indicates that the generated occupancy features are widely distributed across the latent space, reflecting true structural and geometric diversity rather than minor variations of a single collapsed mode.

\subsection{More simulation pipeline details}
\label{sec:sim_pipeline}

\begin{figure}[htb]
    \centering
    \includegraphics[width=\linewidth]{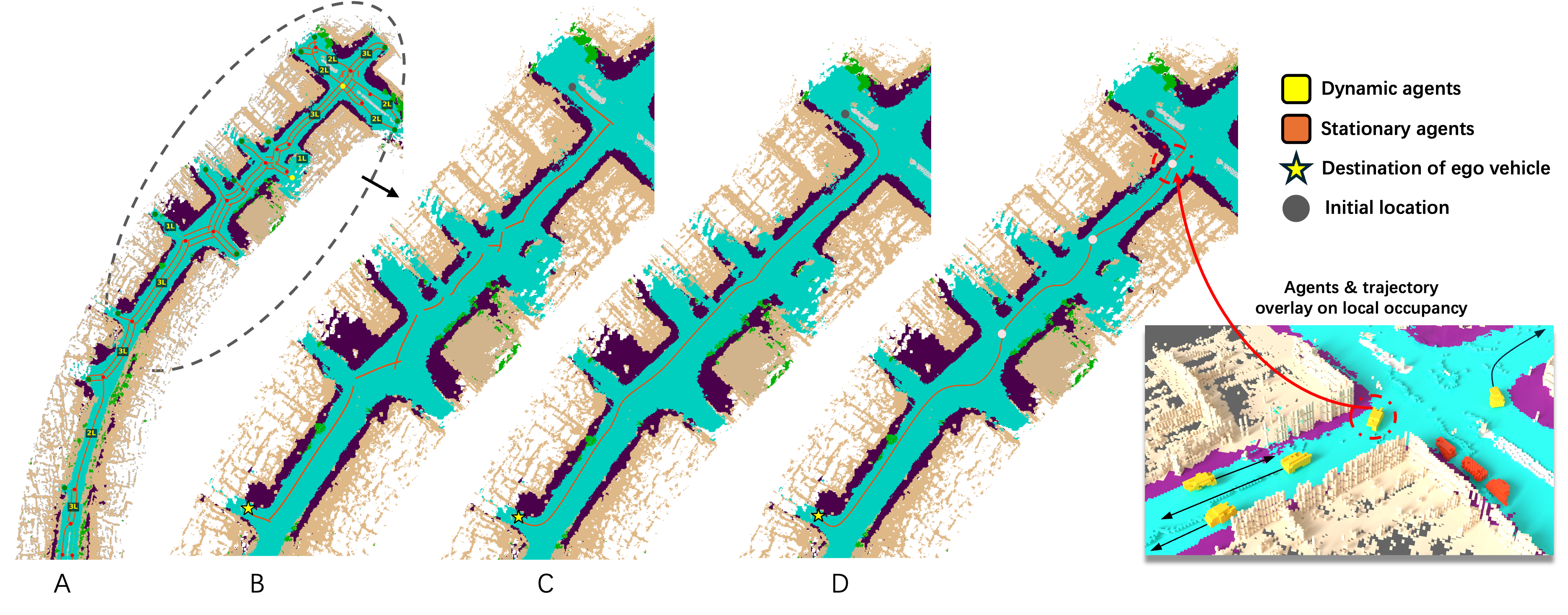}
    \caption{\textbf{The simulation pipeline after static occupancy fusion.}
    \textbf{A}: Part of the lane graph $\mathcal{G}$ extracted from the fused global semantic occupancy $\mathcal{M}_{\text{global}}$, shown overlaid on it. \textbf{B}: All agents, include ego, are instantiated by the same procedure (Layout generator in~\ref{sec:agent}); the ego is distinguished only in that observations are rendered from its pose and itself is not rendered in collected data. Grey dots mark sampled initial poses; each agent independently draws a destination from all possible branch road boundary, and routes to it via A$^\ast$ over the extracted lane graph. \textbf{C}: The routed lane segments are connected and B-Spline smoothed into an initial reference route $\pi_{\text{traj}}$. \textbf{D}: The realized trajectory after closed-loop 2D-IDM execution. Pink dots mark lane switches triggered by agent--agent interaction (Sec.~\ref{sec:2d-idm}); because these depend on the joint state of all agents, the realized trajectories differ across rollouts even when the reference routes coincide. No route or trajectory at any stage is replayed from a recorded log.}
    \label{fig:sim_pipeline}
\end{figure}

In Tab.~\ref{tab:results_trans}, we quantitatively demonstrate an important downstream application of our proposed method: providing data for training of downstream occupancy world model. Here, in Fig.~\ref{fig:sim_pipeline}, we show the detailed process of how ego vehicle was designed to run in our simulation environment. 

Specifically, when simulating the motion behavior of agents (vehicles), we do not distinguish between ego and non-ego vehicles, we just collect the training data with the ego as the center. Therefore, in another words, all agents in the simulation are equivalent and we simply selected a car at random to record the data. Considering the irregularities in the generated occupancy, a closed-loop episode in our setting lasts around 1 min (Tab.~\ref{tab:safety_eval}), whereas Carla can run indefinitely in autopilot mode.

It is worth noting that the simulation data generated by OccSim exhibits a high degree of diversity and non-deterministic trajectories:
\begin{itemize}
    \item W-DiT, which generates static scenes, is multi-modal, which allow us synthesize a large number of occupancy frames with complex road networks.
    \item The multi-modality of layout generator: as the ego vehicle moves forward, agents ahead are generated and rendered into the scene in real time, and each generation of a reasonable position is multimodal.
    \item Each agent's destination is a randomly selected point on the map's boundary, which means that traffic flow behavior can vary even within the same map.
    \item The initial velocity and acceleration of each agent also vary slightly, and when multiple agents move under IDM control, their interactions (avoidance maneuvers) with other agents are completely different in multi rollouts. 
\end{itemize}

\section{Additional ablation experiments}
\label{sec:extra_ablation}

\subsection{Inference speed and condition comparison}
\label{sec:speed}

In the previous section, we demonstrated a comparison between our method and prior approaches in terms of realism. However, it is worth noting that our method also achieves improvements in real-time performance compared to COME. In Tab.~\ref{tab:efficiency_comparison}, we compare the time and FLOPS required for our method to generate a single frame. Our work generate 1 frame per inference, but still more efficient than COME, which generate 6 frames together per inference step.

\begin{table}[htbp]
    \centering
    \caption{Comparison of efficiency and computational cost among different methods. When evaluating the time taken by our method, we distribute the post-processing time evenly across each frame of the sequence generation. Each result was computed independently using an RTX4090. DOME and COME generate 6 frames together, ours is 1 single frame per step, for fair comparison, we keep measure the GFLOPS / inference time needed per single frame.}
    \label{tab:efficiency_comparison}
    \setlength{\tabcolsep}{6pt} 

    \begin{tabular}{lclrrr}
        \toprule
        Method & WM & Input & \#Params & FPS & $\frac{\text{GFLOPS}}{\text{frame}}$ \\
        \midrule
        DOME & \ding{51} & 4 frames sem. occ.\&traj & 444.07M & 5.48 & 8891.98 \\
        COME & \ding{51} & 4 frames sem. occ.\&traj & 204.53M & 2.33 & 41333.17 \\
        InfiniCubes & \ding{53} & 1 frames sem. occ. & 831.75M & 0.02 & 990927.28 \\
        OccSim (Ours) & \ding{51} & 1 frames sem. occ. \& traj & 181.24M & 2.58 & 15570.18 \\
        \bottomrule
    \end{tabular}
\end{table}

\subsection{Validating the lane extraction heuristic against HD maps}
\label{sec:heuristic}
In design of OccSim, we use the heuristic detailed described in Sec.~\ref{sec:lane_topo}, which will later be used as trajectory segments for agents. Here we use the real HD-Map to measure the effectiveness of this approch.

As shown in Tab.~\ref{tab:lane_extraction}, we first rasterize the HD-map drivable-area polygon and feed it to the same heuristic in Alg.~\ref{alg:lane_topology}, which establishes a lower bound on the error that our lane construction incurs regardless of the input quality. We attribute this residual mainly to two properties intrinsic to the construction: the absence of explicit turning connectors at intersections, and the fixed-width assumption of our orthogonal offsetting, which cannot accommodate roads of non-uniform lane width. Substituting GT semantic occupancy for the polygon raises the Chamfer error to 0.74 m, with precision and recall of 0.79 and 0.82. The additional degradation is within one voxel of the 0.4,m grid, and we attribute it to voxel quantization together with the absence of a separate parking-area label in the occupancy taxonomy, which causes parking lanes to be treated as drivable surface.

\begin{table}[t]
\centering
\caption{\textbf{Geometric validation of lane topology extraction against HD maps.}
Both rows run the identical extraction pipeline on the same spatial extent and are evaluated
against the same nuScenes HD-map centerlines; they differ only in the input drivable mask.
The first row uses a rasterized HD-map polygon and thus lower-bounds the error attributable
to our lane construction alone; the excess of the second row is the net cost of operating on
fused occupancy.}
\label{tab:lane_extraction}
\small
\setlength{\tabcolsep}{5pt}
\begin{tabular}{llccc}
\toprule
Drivable mask & Error source
& Chamfer (m) $\downarrow$ & P@1.0m $\uparrow$ & R@1.0m $\uparrow$ \\
\midrule
\rowcolor{gray!15}
HD-map polygon & Definitional (l.b.)
& 0.51 & 0.87 & 0.90\\
GT semantic occupancy & \;\;+ occupancy \& fusion
& 0.74 & 0.79 & 0.82  \\
\bottomrule
\end{tabular}
\end{table}

\subsection{More details in few steps sampling \& single step rollout stride}
\label{sec:nfe}

\begin{wrapfigure}[16]{r}{0.5\textwidth}
\centering
\vspace{-\intextsep}
\includegraphics[width=0.48\textwidth]{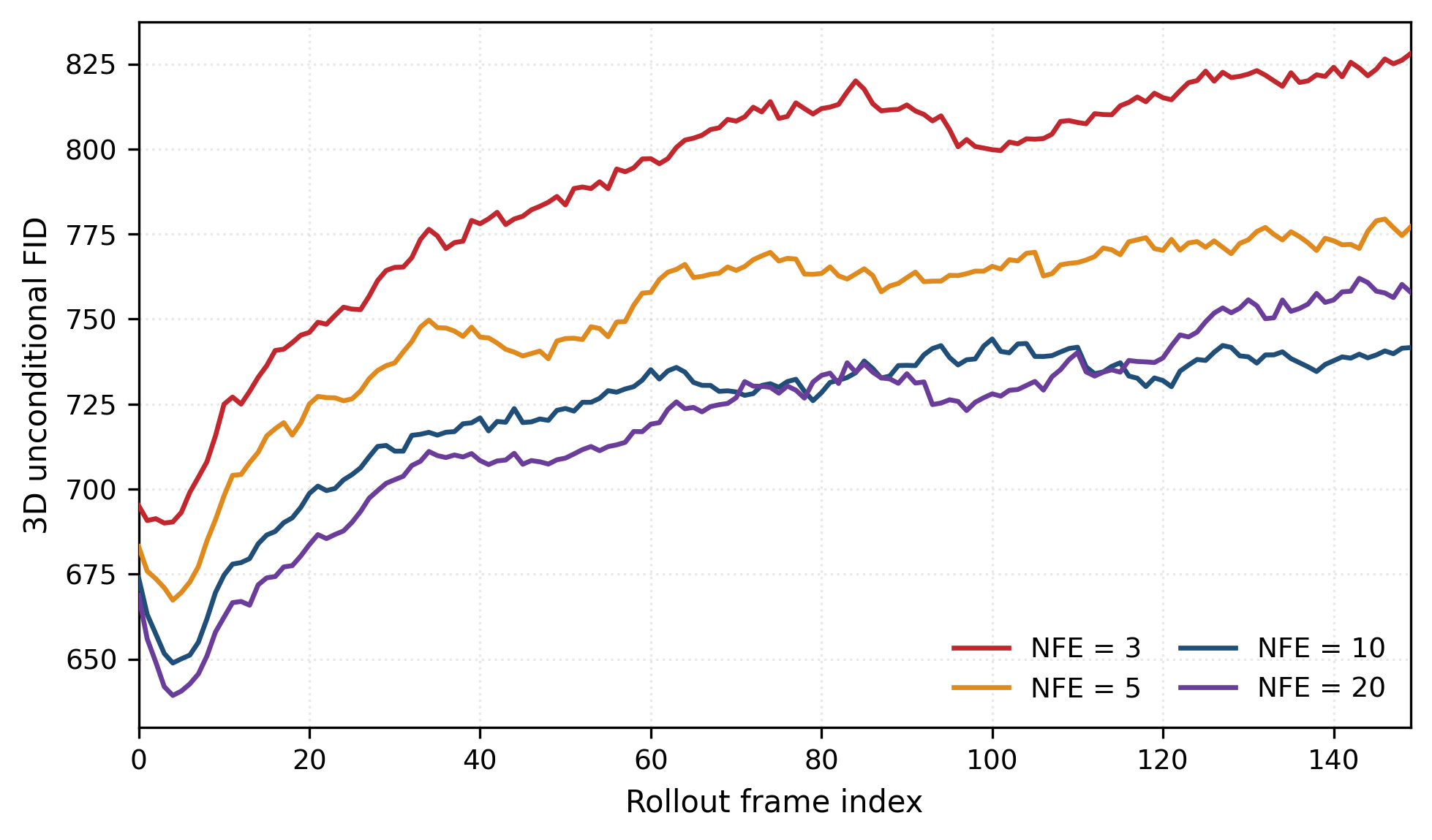}
\caption{\textbf{Effect of sampling steps (NFE) on rollout quality.}
Per-frame 3D unconditional FID against the training-set marginal, over 150-frame
rollouts on the UniScenes AE latent (3\,m per step, $\text{cfg}=1.1$).
Fidelity does not improve beyond NFE${=}10$, which we adopt as the operating point.}
\label{fig:nfe_ablation}
\end{wrapfigure}

\begin{figure}[htb]
    \centering
    \includegraphics[width=\linewidth]{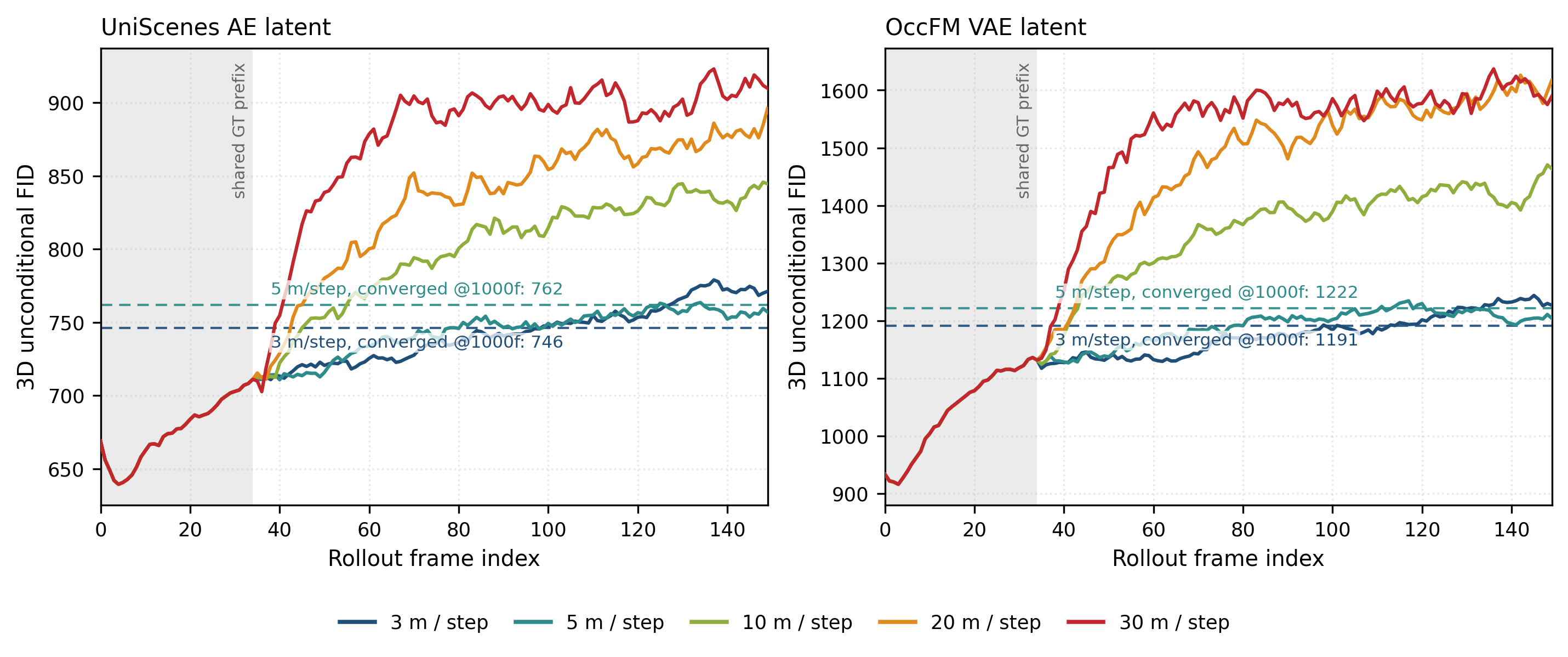}
    \caption{Fidelity of generated frames over time under different moving steps. We use the ground truth trajectories and poses at first 35 frames, concatenated with straight trajectory moving with different stride in single step. We use 3m / step for all of the experiments mentioned in the main text.} 
    \label{fig:step_size}
\end{figure}

For NFE step, as shown in Fig.~\ref{fig:nfe_ablation}a, NFE=10 get the best and stablest results in terms of fidelity. For the rollout stride, we use straight trajectory in Fig.~\ref{fig:traj_vis} to test fidelity of rollouts under different single step stride, from 3m to 30m. As shown in figure~\ref{fig:step_size}, we find that 3 meters per step achieve the best results. 

We do not attributes this to out-of-distribution(OOD) in update size: During training, $\mathbf{M}_{\mathrm{rand}}$ is composed of one to three random rectangular blocks whose spatial extent spans roughly 8--56\,m (overall 10\%-50\% of the full scene); a 30\,m forward band is therefore not OOD in size, and the model is in principle capable of completing an unknown region at that scale in a single pass.

We instead attribute the degradation to the autoregressive structure itself: since W-DiT re-synthesizes the complete frame at every step, a given spatial location is refined $\lceil \mathrm{FOV}/\Delta d \rceil$ times over its visible lifetime, and this count falls as the stride $\Delta d$ grows (FOV: Field of View, $80m \times 80m$ here). Prior outpainting-based methods like InfiniCube \citep{lu2025infinicube} obtain an \textbf{equivalent effect} through repeated intra-block resampling \citep{lugmayr2022repaint}, at the cost of a $64\times$ larger per-frame compute budget (Tab.~\ref{tab:efficiency_comparison}); our stride offers a cheaper form of the same fidelity--throughput trade-off, and the ablation shows that the returns are already saturated at 3--5\,m.

\subsection{Spherical readout: a norm-invariant semantic head}
\label{sec:sphere}

\begin{figure}[h]
\centering
\begin{minipage}[t]{0.48\textwidth}
\vspace{0pt}
\centering
\includegraphics[width=\linewidth]{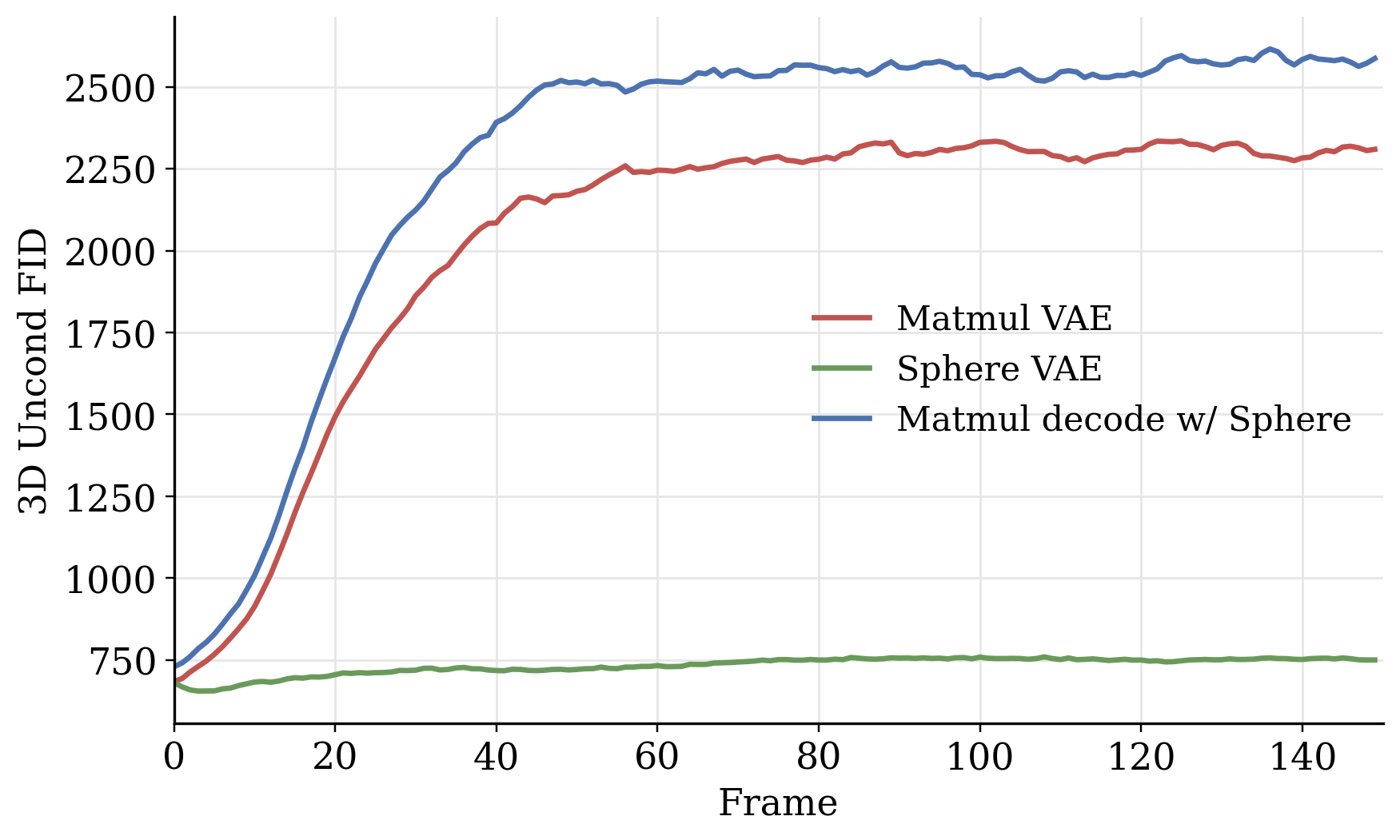}
\captionof{figure}{Long-horizon 3D unconditional FID in the UniScenesAE latent space. Standard matmul decoding degrades rapidly as latent energy decays over rollout, while norm-invariant sphere decoding stays stable across all 150 frames.}
\label{fig:sphere_decode_fid}
\end{minipage}
\hfill
\begin{minipage}[t]{0.48\textwidth}
\vspace{0pt}
\centering
\begin{algorithm}[H]
\caption{Norm-Invariant Sphere Decoding}
\label{alg:sphere_decode}
\begin{algorithmic}[1]
\Require Decoded feature map $\mathbf{F}\in\mathbb{R}^{B\times H\times W\times Z\times C}$, semantic template $\mathbf{W}\in\mathbb{R}^{C\times K}$, scale $s$
\Ensure Class logits $\mathbf{L}\in\mathbb{R}^{B\times K\times H\times W\times Z}$
\If{method $=$ \texttt{sphere}}
    \State $\hat{\mathbf{F}} \gets \mathbf{F}/\|\mathbf{F}\|_2$ \Comment{normalize over $C$ (dim $=-1$)}
    \State $\hat{\mathbf{W}} \gets \mathbf{W}/\|\mathbf{W}\|_2$ \Comment{normalize over $C$ (dim $=1$)}
    \State $\mathbf{S} \gets \hat{\mathbf{F}}\,\hat{\mathbf{W}}$ \Comment{cosine similarity}
    \State $\mathbf{S} \gets s\cdot\mathbf{S}$ \Comment{fixed temperature $s$}
\Else
    \State $\mathbf{S} \gets \mathbf{F}\,\mathbf{W}$ \Comment{standard matmul; logit $\propto\|\mathbf{F}\|$}
\EndIf
\State $\mathbf{L} \gets \text{permute}(\mathbf{S})$
\State \Return $\mathbf{L}$
\end{algorithmic}
\end{algorithm}
\end{minipage}
\end{figure}

In Tab.~\ref{tab:ablation}, we analyzed several factors that influence performance in W-DiT, including the loss function and architectural design. However, we found that the special designed representation readout method is also essential for stable, long-term rollouts. Specifically, when training the VAE for occupancy latent compression, we revise readout method from standard matmul to sphere decode, as shown in Alg.~\ref{alg:sphere_decode}.

\begin{table}[h]
\centering
\begin{minipage}[t]{0.48\textwidth}
\vspace{0pt}
\centering
\caption{Single-frame VAE encode--decode reconstruction on Occ3D-nuScenes (C18). Sphere decoding has almost the same IoU and mIoU.}
\label{tab:sphere_decode}
\setlength{\tabcolsep}{6pt}
\renewcommand{\arraystretch}{1.2}
\begin{tabular}{lcc}
\toprule
\textbf{VAE Decode Method} & \textbf{IoU} & \textbf{mIoU} \\
\midrule
Matmul & 74.66 & 85.99 \\
Sphere & \textbf{75.15} & \textbf{86.27} \\
\bottomrule
\end{tabular}
\end{minipage}
\hfill
\begin{minipage}[t]{0.48\textwidth}
\vspace{0pt}
\centering
\caption{Geometric analyze of the semantic template $\mathbf{W}$ before (Matmul) and after (Sphere) $\ell_2$ normalization. CV.: coefficient of variation. PC.: principal component. MACS: Mean Absolute Cosine Similarity.}
\label{tab:template_geometry}
\setlength{\tabcolsep}{5pt}
\renewcommand{\arraystretch}{1.2}
\begin{tabular}{lcc}
\toprule
\textbf{Template statistic} & \textbf{Matmul} & \textbf{Sphere} \\
\midrule
Avg.\ length & 2.801 & 1.000 \\
Length CV. ($\downarrow$) & 0.4385 & \textbf{0} \\
Top-3 PC. ratio & 58.73\% & 49.89\% \\
MACS. ($\downarrow$) & 0.2324 & \textbf{0.2207} \\
\bottomrule
\end{tabular}
\end{minipage}
\end{table}

As shown in Fig.~\ref{fig:sphere_decode_fid} and Tab.~\ref{tab:sphere_decode}, the sphere readout has almost the same reconstruction performance as original matmul-based method, but significantly help the rollout stability. Furthermore, when we applied the sphere readout only during testing to VAE representations trained using the matmul readout, as shown by the blue line in Fig.~\ref{fig:sphere_decode_fid}, the results were actually worse. This experiment decoupled the decoder design itself from its effect on constraining the latent manifold during VAE training, demonstrating that the sphere readout implicitly reshapes the geometry of the latent manifold, and that this constraint ensures stability. 

Treating each generated latent frame $z \in \mathbb{R}^{50 \times 50 \times 16}$ as 2,500 tokens of length 16, the latents from approximately 28,000 samples in the training set form a 16-dimensional space comprising approximately $7 \times 10^{7}$ tokens. In this space, we evaluate angular(which isolates directional drift) and radial distance to reflect the OOD of the latents under VAEs with different readout schemes. For rollout frame $z_i^{(t)}$ at frame t, the angular distance is its k-th nearest-neighbor $\pi_{k}(i)$ distance in ground-truth latent $z_{gt}$ on the unit sphere, defined below, here k=10. Radial distance is the angular distance without unit sphere constrain. Each is a median over the 2,500 tokens per frame, then across sequences (band: 25–75th pct.).

$$\label{eq:ood}
d^{\text{ang}}_t = \operatorname*{median}_i \big\| \hat{z}_i^{(t)} - \hat{z}_{\text{gt},\,\pi_k(i)} \big\|_2,
\qquad \hat{z} = z / \|z\|_2 .$$

\begin{figure}[t]
\begin{minipage}[t]{0.44\textwidth}
    \centering
    \includegraphics[width=\textwidth]{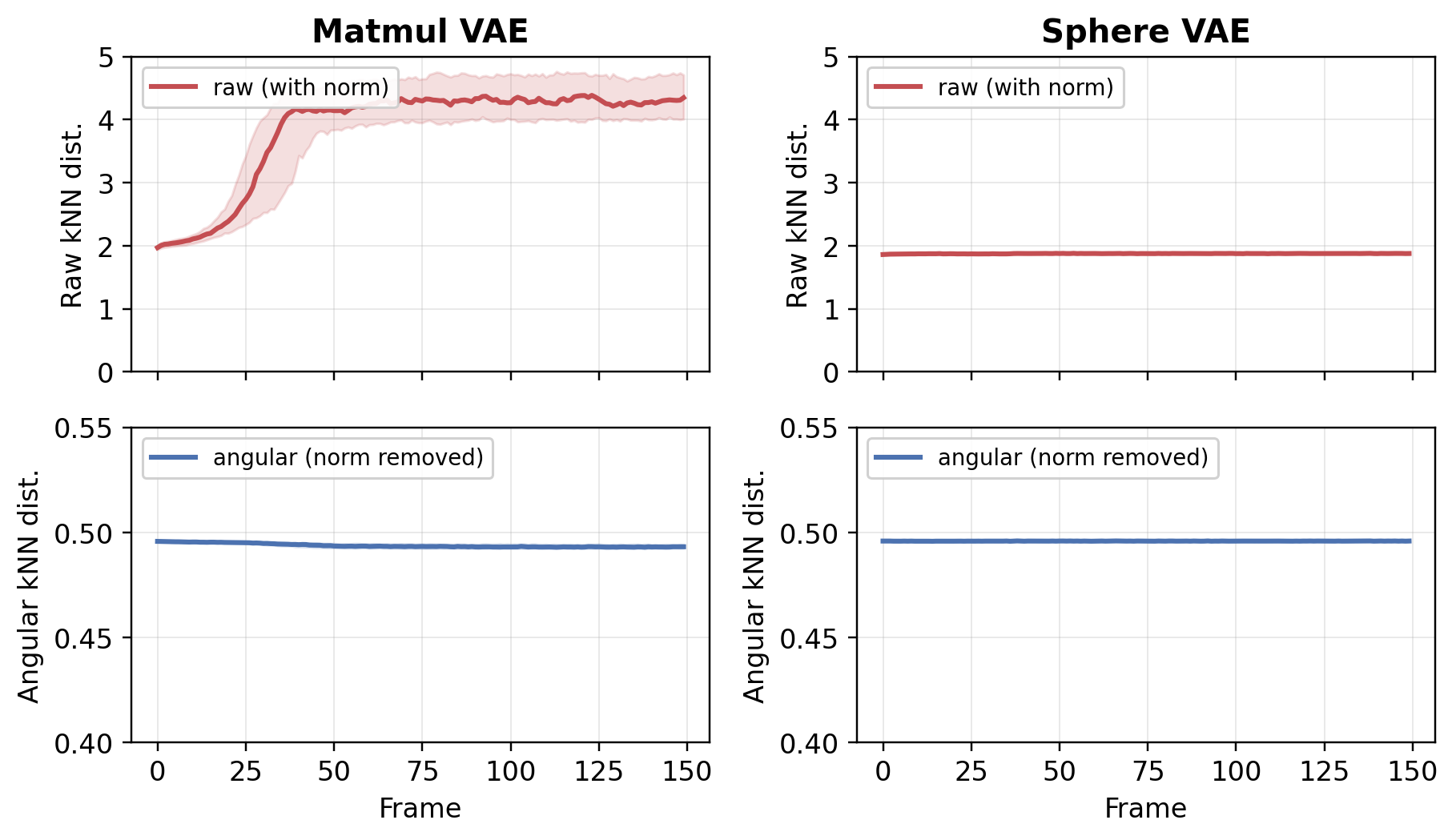}
    \caption{Per-token latent OOD within each VAE's own space: raw kNN (with norm) inflates only under matmul, while angular kNN (norm removed) stays flat for both.}
    \label{fig:ood}
\end{minipage}
\hfill
\begin{minipage}[t]{0.54\textwidth}
    \centering
    \includegraphics[width=\textwidth]{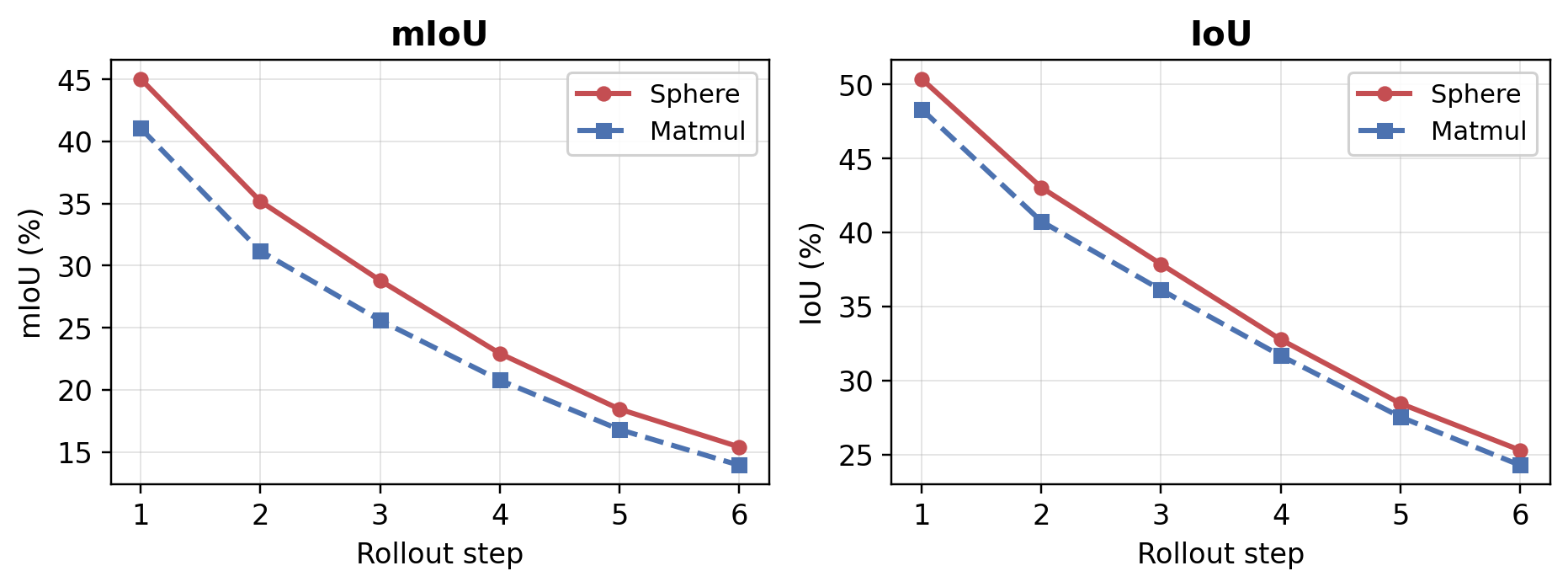}
    \caption{Downstream forecasting mIoU / IoU across rollout steps. Retraining the VAE with the sphere readout improves both metrics over the matmul baseline with almost no extra cost.}
    \label{fig:sphere_downstream}
\end{minipage}
\end{figure}

The results are shown in Fig.~\ref{fig:ood}. Under the matmul VAE, the raw kNN distance inflates by $\sim2.2\times$over the rollout while the angular distance stays essentially flat ($\sim0.99\times$); the sphere VAE remains stable in both. Since raw and angular kNN differ only in whether the token norm is retained, this shows the matmul latent drifts \emph{purely radially}(norm inflation), not by directional change. Although the absolute distances of the two sub-figures are not directly comparable (they live in different latent spaces), the within-space trends consistently indicate that the sphere readout stabilizes the rollout by constraining the norm geometry of the latent space, mirroring the static norm imbalance in Tab.~\ref{tab:template_geometry}. Further analysis in Tab.~\ref{tab:template_geometry} show that normalization removes inter-class norm imbalance (length CV $\to 0$) in template $\mathbf{W}$ while leaving inter-class \emph{directions} almost untouched (MACS nearly unchanged), and flattens the singular-value spectrum toward isotropy.

Finally, we further verify that the sphere readout is not specific to our architecture: revising OccFM\citep{liu2025towards}'s VAE with the sphere readout from the matmul based one improves its downstream forecasting mIoU by several points, indicating the gain comes from the geometric regularity it imposes on the latent space rather than any model-specific design. As shown in Fig.~\ref{fig:sphere_downstream}, the sphere readout thus offers an almost free improvement (+2.74\% absolutely, +11\% relatively): no extra parameters cost, no reconstruction trade-off, yet a consistent downstream gain. 

\subsection{Complementing Real Logs with OccSim Data}
\label{sec:augmentation}

\begin{table}[t]
\centering
\caption{OccWorld trained on real UniOcc-Waymo logs (Uni-W), OccSim data ($5\times$ volume, as in Tab.~\ref{tab:results_trans}), their mixture (OccSim data amounting to 30\% of Uni-W), and CARLA (Uni-C), evaluated zero-shot on the UniOcc-nuScenes validation set.}
\label{tab:real_mix}
\begin{tabular}{lcccccc}
\toprule
\multirow{2}{*}{Train Data} & \multicolumn{3}{c}{mIoU $\uparrow$} & \multicolumn{3}{c}{IoU $\uparrow$} \\
\cmidrule(lr){2-4} \cmidrule(lr){5-7}
 & 1s & 2s & 3s & 1s & 2s & 3s \\
\midrule
Uni-W & 23.47 & 18.11 & 15.80 & 27.35 & 20.86 & 17.63 \\
Uni-W + OccSim & 24.95 & 19.71 & 15.08 & 29.73 & 22.11 & 18.49 \\
OccSim & 19.54 & 12.31 &  9.08 & 23.87 & 16.92 & 13.47 \\

CARLA & 11.79  & 8.35  & 6.75 & 18.31 & 14.09 & 12.03 \\

\bottomrule
\end{tabular}
\end{table}

Tab.~\ref{tab:real_mix} compares OccWorld trained on real UniOcc-Waymo logs (Uni-W), on OccSim data alone, and on their mixture, all evaluated zero-shot on UniOcc-nuScenes. Trained alone, OccSim data remains below real logs, which is expected: real logs are the realism upper bound of the data OccSim learns from. It nonetheless outperforms the asset-based CARLA data by a large margin. The value of OccSim lies in what logs cannot offer: a log is a fixed recording, whereas OccSim generates new road topologies and traffic layouts on demand and supports closed-loop interaction. Adding OccSim data amounting to 30\% of Uni-W improves the average IoU over 1--3s by 6.8\% relative (21.95 to 23.44) and the average mIoU by 4.1\% (19.13 to 19.91), with gains at every horizon except 3s mIoU, which decreases slightly (15.80 to 15.08). OccSim data thus complements real logs without additional annotation. As in Tab.~\ref{tab:results_trans}, all metrics here are computed open-loop against logged futures (see limitation 5 in Sec.~\ref{sec:limitation}).

\section{Additional visualization}

We provide more visualizations for details of our work here. More information like dynamic traffic simulated can be found in the video in the supplementary materials.

\subsection{Details of trajectory shapes used in Fig.~\ref{fig:cond_3d}}
\label{sec:appendix_trajectories}

\begin{figure}[htb]
    \centering
    \includegraphics[width=0.85\linewidth]{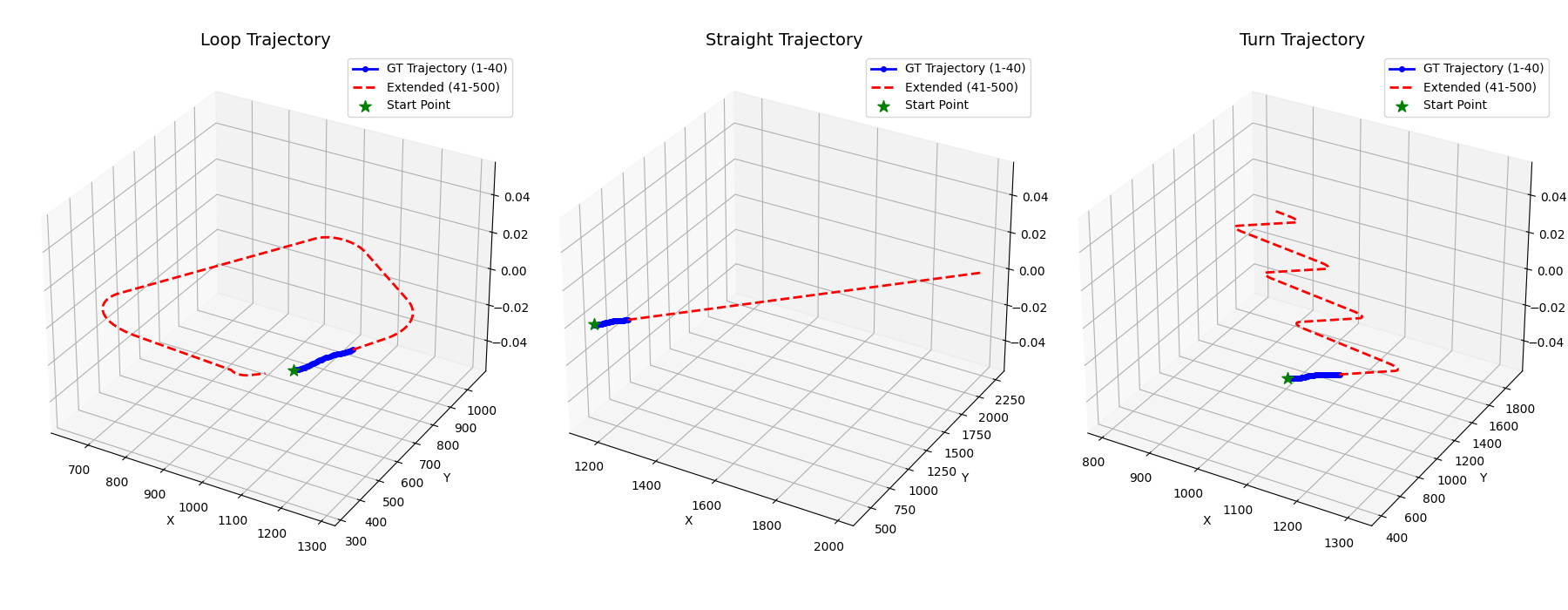}
    \caption{Visualization of trajectory shapes (Loop, Straight, and Turn) evaluated in Fig. \ref{fig:cond_3d}. The green star indicates the starting point. The blue solid line denotes the initial ground-truth (GT) trajectory snippet (frames 1–40) used as the condition prompt, while the red dashed line represents the extended trajectory explicitly designed to guide the long-horizon generation. For visual clarity the displayed trajectories are truncated at 500 frames, whereas our actual synthesis spans up to 3000 frames.}
    \label{fig:traj_vis}
\end{figure}

To systematically evaluate the robustness of our model in long-horizon generation across diverse driving behaviors, we configure three distinct spatial trajectory profiles: Loop, Straight, and Turn. As illustrated in Fig.~\ref{fig:traj_vis}, the trajectory generation process conditioned on is consist of  the initial ground-truth (GT) trajectory (highlighted in blue, encompassing frames 1–40) and extended trajectory with 3 modes (red lines). 

\subsection{Comparison with other Occ. World Models in long-horizon rollouts}
\begin{figure}[htb]
    \centering
    \includegraphics[width=\linewidth]{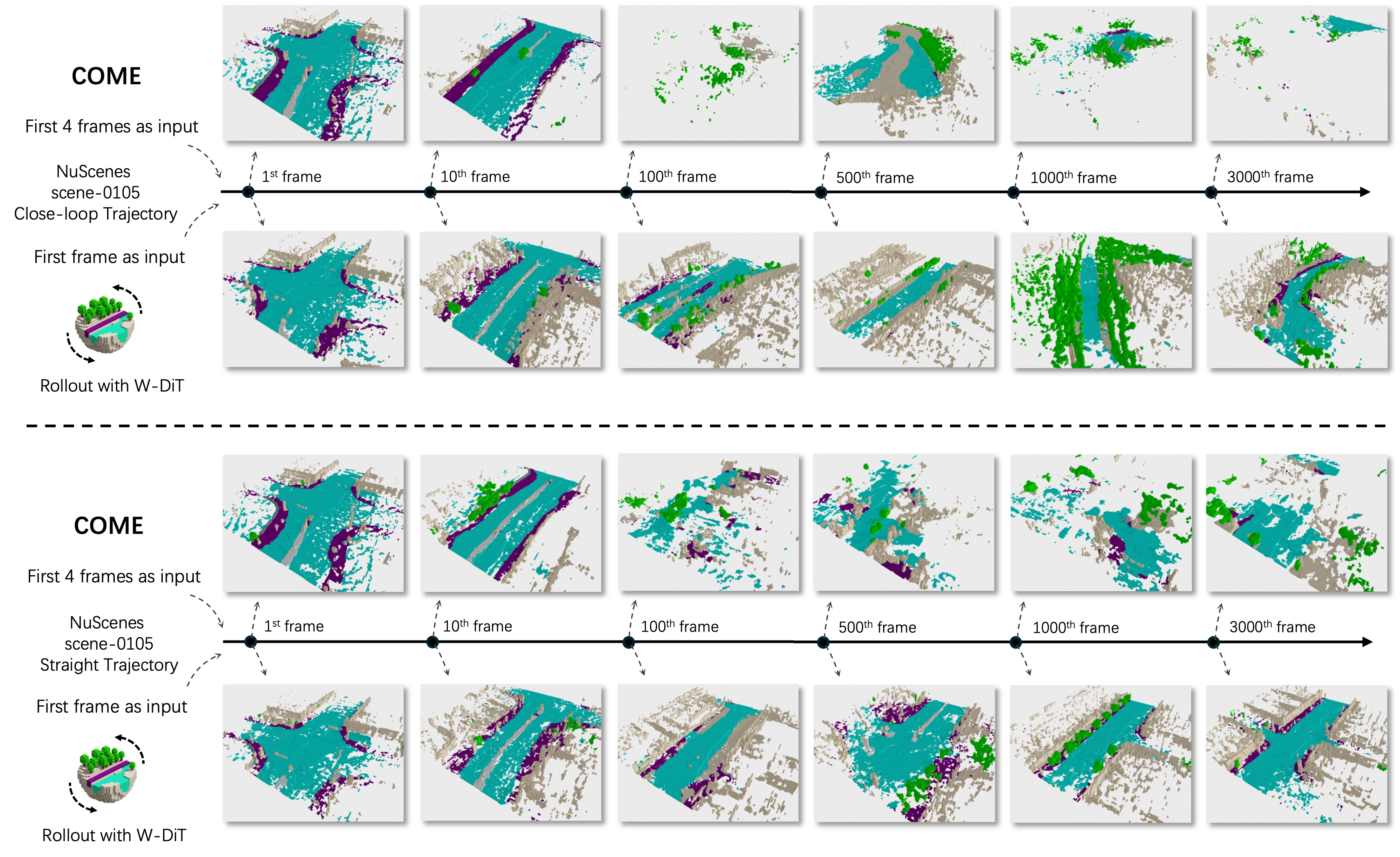}
    \caption{Qualitative comparison of two rollouts for two models: previous SOTA occupancy world model (first and third rows), and our model (second and fourth rows) at different horizons. Ours exhibits long-horizon stability under different trajectory input.}
    \label{fig:w-dit_over_time}
\end{figure}

From Fig.~\ref{fig:w-dit_over_time}, we can clearly observe that our proposed W-DiT significantly extends the duration of stable rollout. Even after 3000 frames (1500s, approximately 25 minutes with 2Hz setting in Occ3d-NuSc), the generated occupancy remains stable. The top half of Fig.~\ref{fig:w-dit_over_time} shows the output when the loop trajectory from Fig.~\ref{fig:traj_vis} is used as input, while the bottom half shows the output when the straight-line trajectory from Fig.~\ref{fig:traj_vis} is used as input. 

\subsection{Comparison with out-painting based methods}
\label{sec:comp_outpaint}

\begin{figure}[h]
    \centering
    \includegraphics[width=\linewidth]{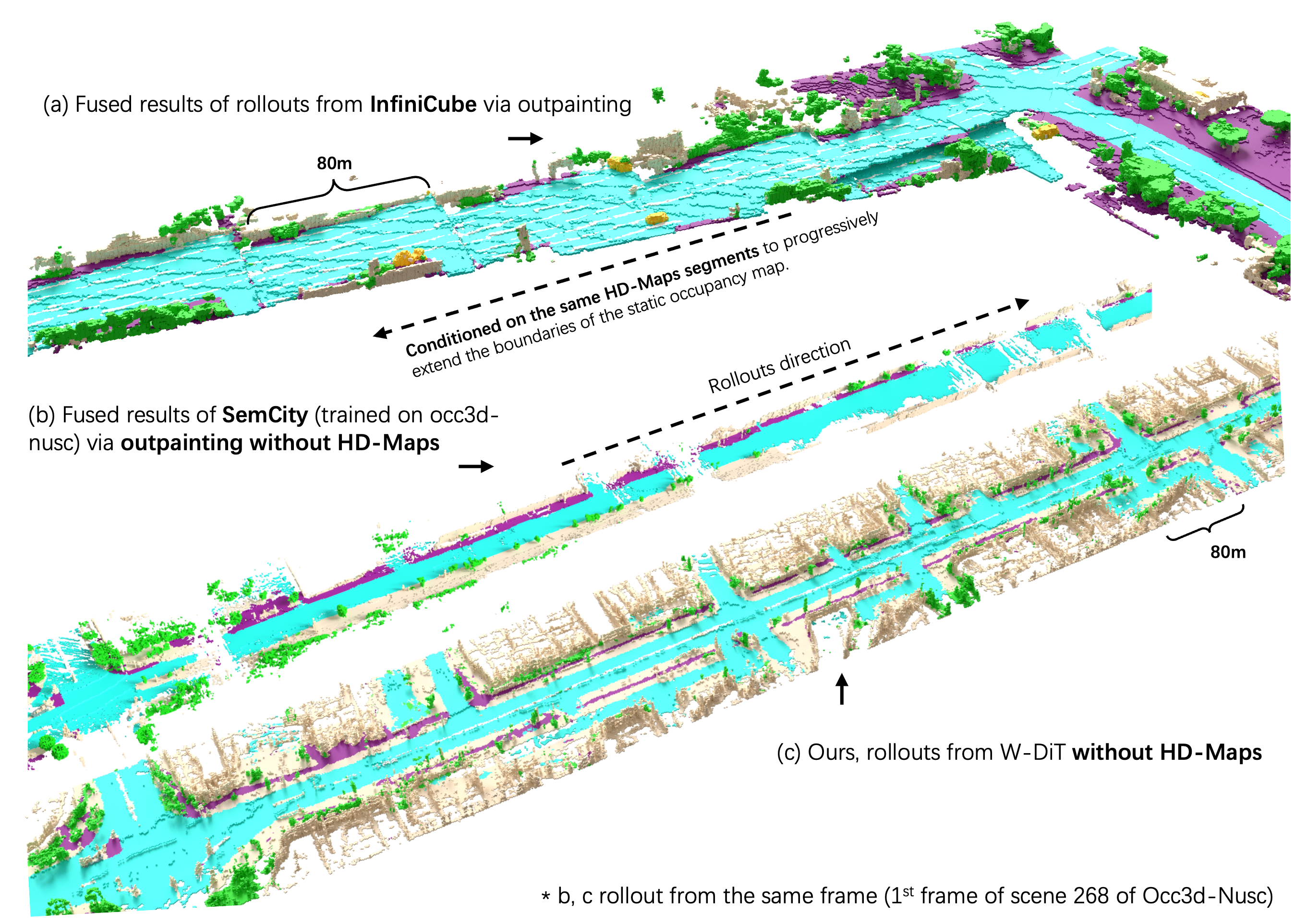}
    \caption{Visualization of static occupancy generated from InfiniCube, SemCity and our approach (from top to down). For InfiniCube, we reuse the boundary HD-map to continuously extend the map along a straight line. Likewise, for SemCity and our method, we extend the scene in the same direction.}
    \label{fig:comp_non_wm}
\end{figure}

World models are indeed one of the technologies used to generate boundary-free driving scenes; however, outpainting-based methods can achieve the same in the context of static occupancy. Therefore, in this section, we aim to qualitatively demonstrate that our proposed method outperforms previous outpainting-based approaches. In Fig.~\ref{fig:comp_non_wm}, we compare our model against SemCity~\citep{lee2024semcity} and InfiniCube~\citep{lu2025infinicube}, both of which generate future scenes via outpainting. Notably, InfiniCube additionally requires an HD-map aligned with the region to be generated.

As shown in the Fig.~\ref{fig:comp_non_wm}, the outpainting-based InfiniCube not only requires extra HD-map annotations but also suffers from severely limited diversity, exhibiting only minor variations in the foreground across different segments. In contrast, among the HD-map-free approaches, SemCity's results are markedly inferior to ours in realism, diversity, and continuity.

\subsection{Ultra long rollout from W-DiT: 100+km city scale generation with real world trajectory.}
\label{sec:100km}
\begin{figure}[h]
    \centering
    \includegraphics[width=\linewidth]{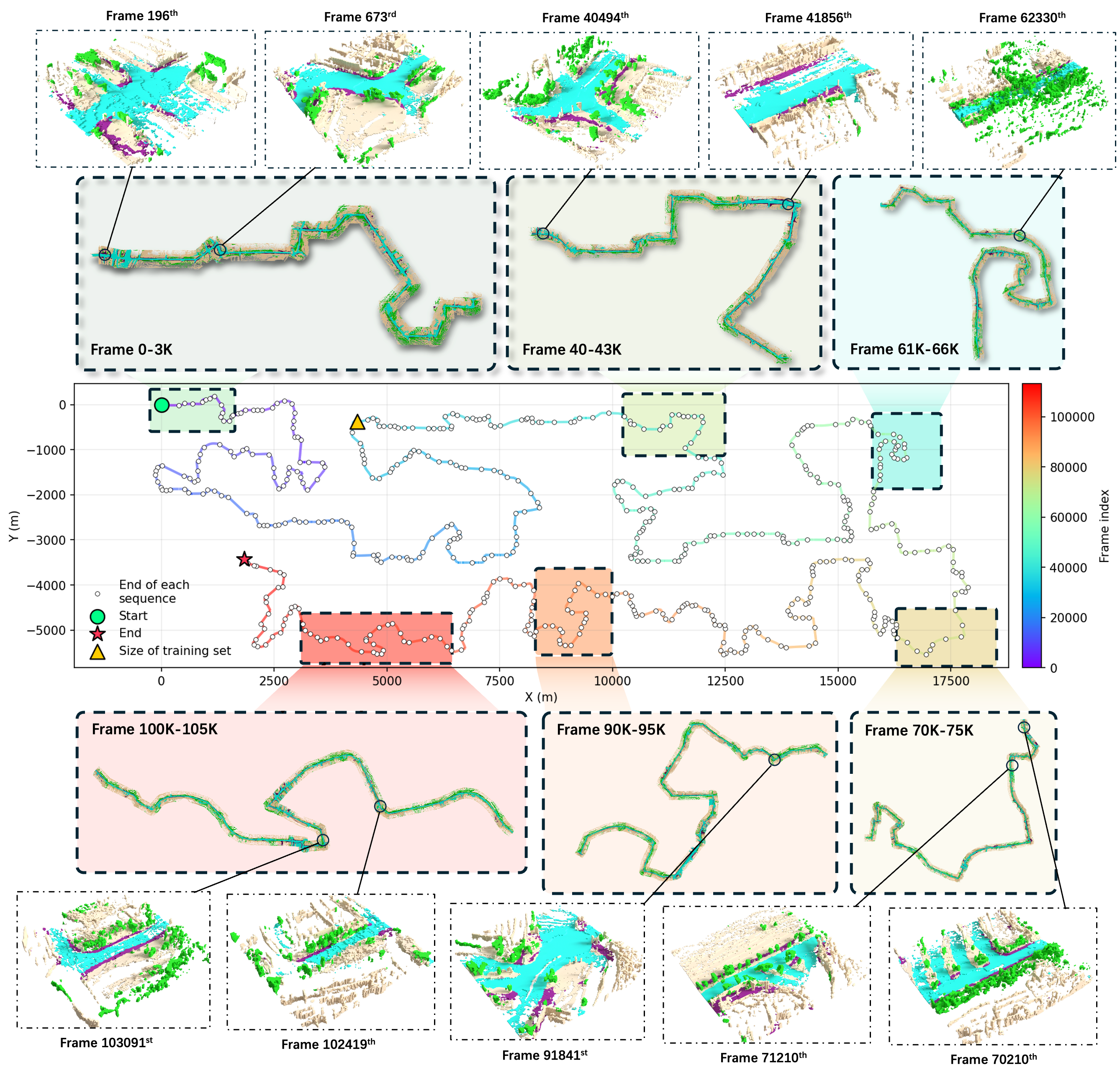}
    \caption{The full 100+ km trajectory (113321 frames) concatenated from Occ3D-Waymo. The connecting angle between every two pose sequences is set manually, and the full trajectory is smoothed with a B-spline after concatenation. This experiment uses the raw step size from real Waymo trajectories (approximately 0.9 m per frame).}
    \label{fig:traj_100km}
\end{figure}

In Fig.~\ref{fig:cond_3d}, we evaluate the fidelity of model rollout, our proposed method remained stable over a time scale of 1,000 frames. Furthermore, we used a 3,000-frame rollout to construct a static map and collected data based on this map to conduct the experiments presented in Tab.~\ref{tab:results_trans}. Here we push the length of generation to 100x longer, build a city-scale occupancy in Fig.~\ref{fig:traj_100km}, to further show the stability of W-DiT.

By manually concatenating all real-world trajectories from the occ3d-waymo training, we obtained driving trajectories from real drivers totaling over 100 km in length (shown at middle of Fig.~\ref{fig:traj_100km}). With this as the user-steered input of our model trained purely on occ3d-nuscenes, we further expanded the rollout to 110K frames, resulting in city-level occupancy data spanning over 100 km.

Specifically, after fully generating these approximately 110K frames, we fused them using Alg.~\ref{alg:keyframe_fusion} and presented the fusion results for a portion of the subsequences. Furthermore, we randomly visualized 10 frames from this sequence. As can be seen, driven by a large number of real-world trajectories, our method is able to generate highly diverse and realistic road structures, which provides a foundation for subsequent data collection and training in downstream tasks. As far as we know, this is the largest scale of driving scenario generation. 

\subsection{Multi-modal diversity of W-DiT: fix-seed multi-time rollout results}

\begin{figure}[h]
    \centering
    \begin{subfigure}[t]{\textwidth}
        \centering
        \includegraphics[width=\textwidth]{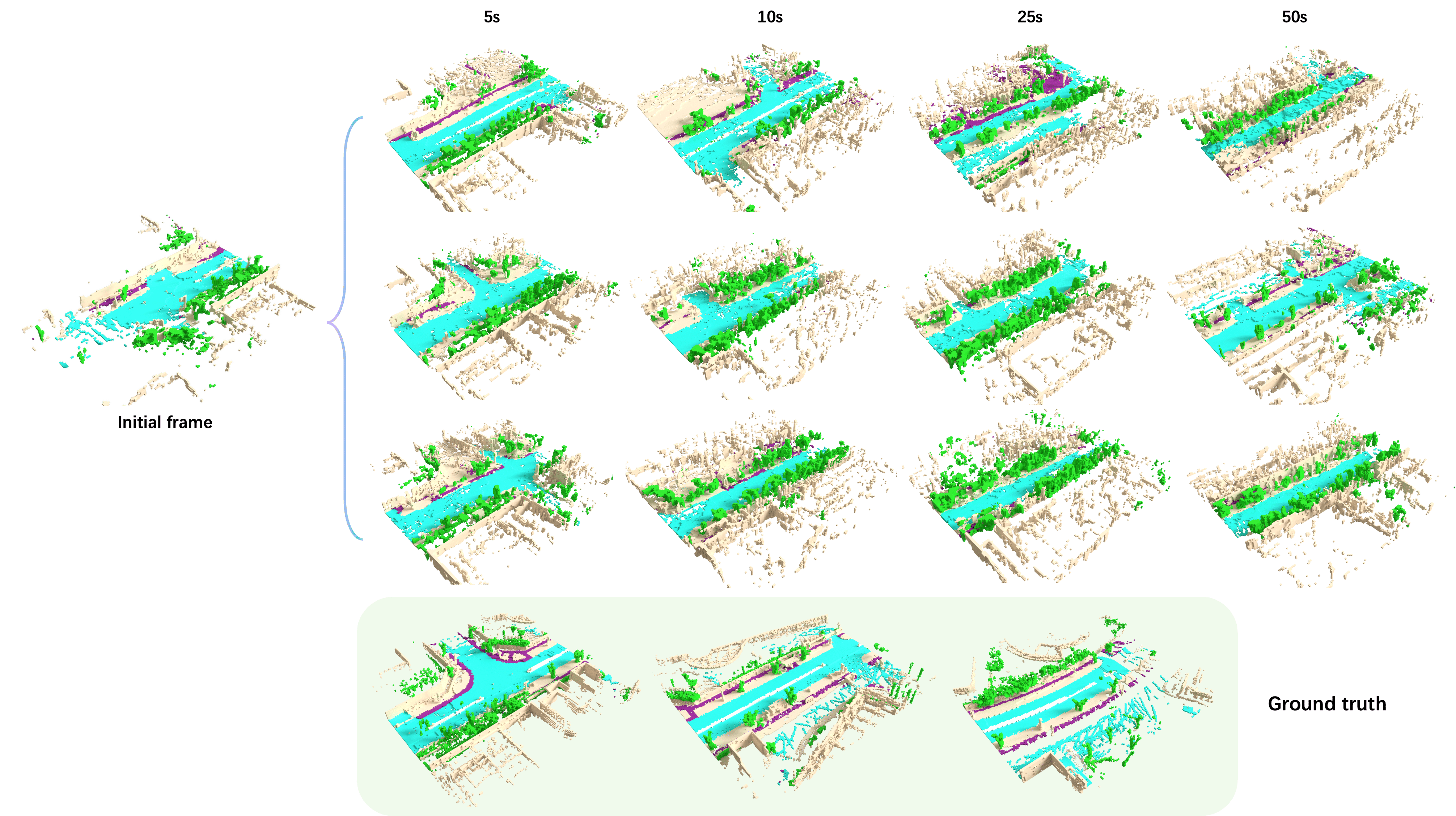}
        \caption{Scene 0039}
        \label{fig:multiseed_scene0039}
    \end{subfigure}
    
    \hfill
    \begin{subfigure}[t]{\textwidth}
        \centering
        \includegraphics[width=\textwidth]{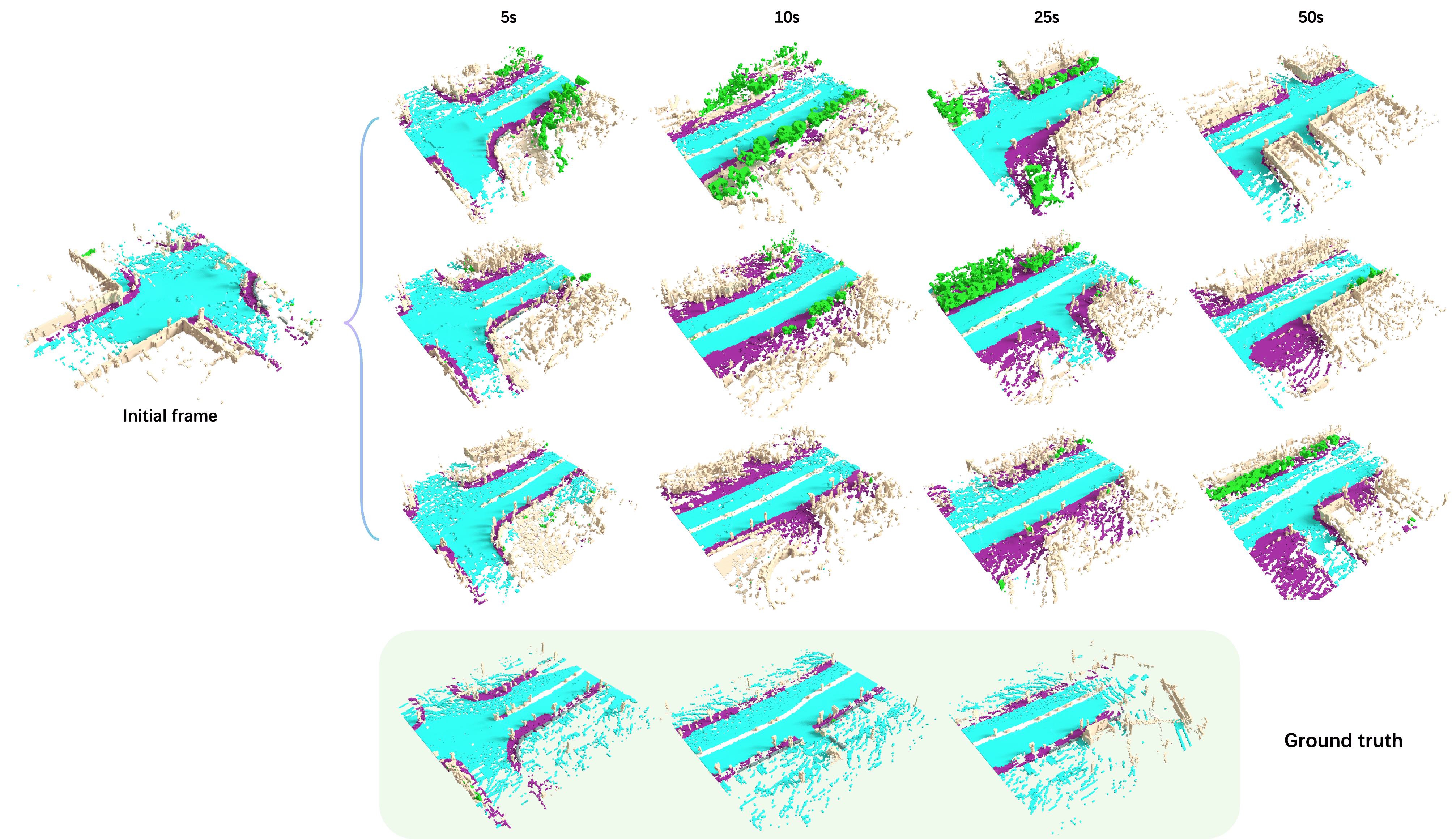}
        \caption{Scene 0105}
        \label{fig:multiseed_scene0105}
    \end{subfigure}
    \caption{Visualization of W-DiT's long-horizon rollouts under different random seeds: given the same input frame (the first frame of scene-039 and scene-105 from Occ3D-nuScenes) and an identical straight-line trajectory. The three rows correspond to seeds 999/2026/3407, respectively.  We also show the comparison with GT sequences at 5s, 10s and 20s here, to further illustrate that our method generate diverse future map, instead of just memorize the ground truth. }
    \label{fig:multiseed_rollout_nusc}
\end{figure}

In the main text, we quantitatively demonstrate the SoTA diversity of our W-DiT-based model rollouts using the Vendi Score. Here, in Figure~\ref{fig:multiseed_rollout_nusc}, we qualitatively illustrate our model's ability to capture the multi-modality of the future when only the input noise (seed) differs.  After 50s (100 frames, we use 6m/s speed to generate trajectory) rollout, the branch road topology, vegetation / man-made presents obvious difference.

\subsection{Visualization of chaos baseline constructed}
\label{sec:appendix_chaos}
\begin{figure}[h]
    \centering
    \includegraphics[width=\linewidth]{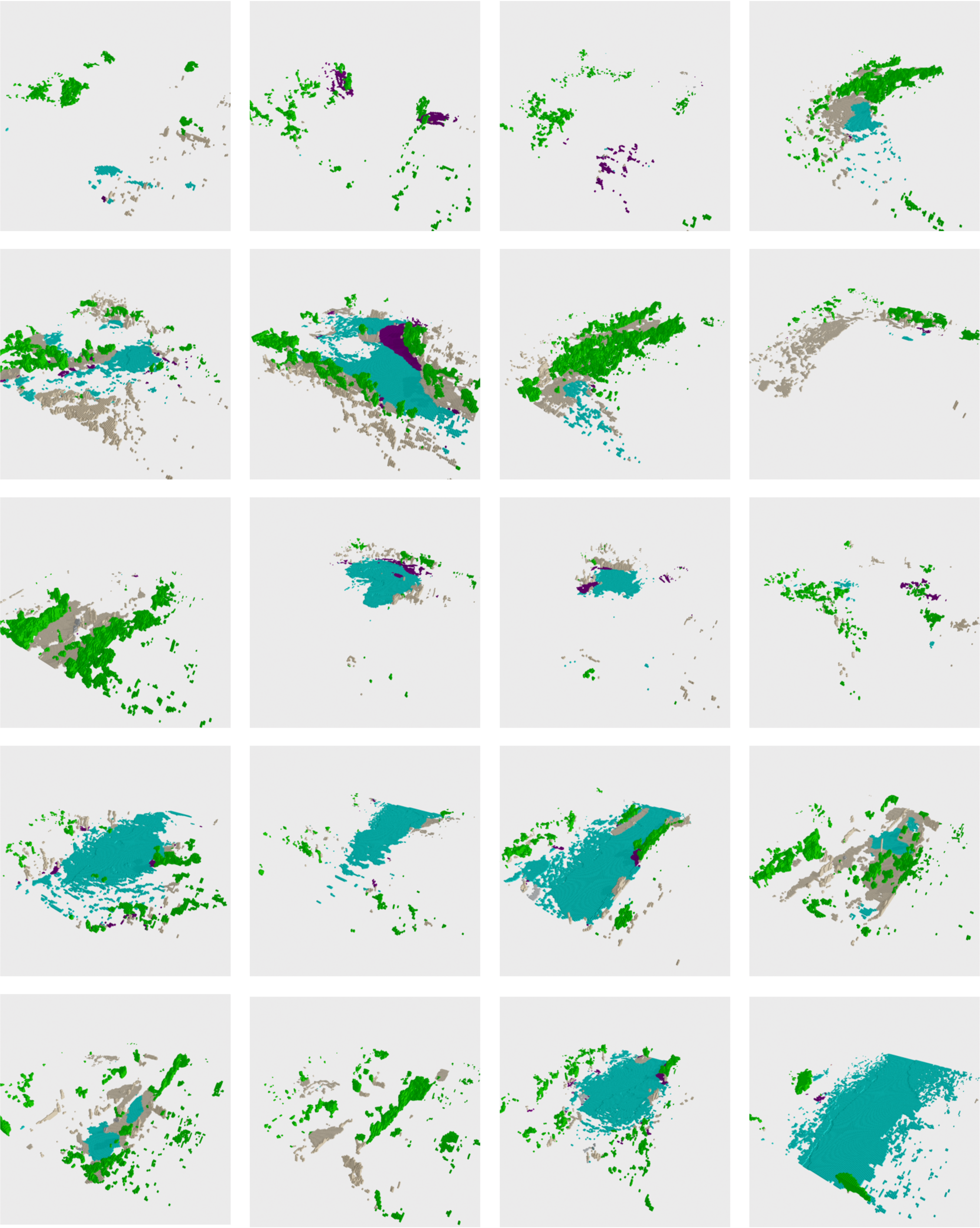}
    \caption{Visualization of frames from the ``chaos'' baseline. We randomly visualize 20 frames from the 1000 samples pool. The semantic categories represented by colors here are consistent with those in Fig.~\ref{fig:with_agent}. }
    \label{fig:chaos}
\end{figure}

As shown in Fig.~\ref{fig:chaos}, the constructed ``chaos'' baseline primarily comprises severely fragmented road segments, topologically disconnected layouts, and large-scale empty regions. Consequently, as we mentioned in Fig.~\ref{fig:cond_3d}, any model yielding FID/KID scores above this baseline threshold can be deemed completely incapable of synthesizing large-scale, valid road networks. 

\section{Algorithm details}
\label{sec:appendix_algorithms}

\subsection{Heuristic for Single frame fusion.} After we obtained the raw forecast observation $\{\mathcal{O}_i\}_{i}^{K}$ from the W-DiT based world model, we than use the following algorithm~\ref{alg:keyframe_fusion} to fuse all frames to $\mathcal{M}_{\text{global}}$. 

\begin{algorithm}[h]
\caption{Two-pass Keyframe-based Occupancy Fusion}
\label{alg:keyframe_fusion}
\begin{algorithmic}[1]

\Require Local predictions $\{\hat{\mathcal{O}}_t\}_{t=1}^T$, Ego-poses $\{\mathbf{T}_t\}_{t=1}^T$, max step $d_{\max}$, vote threshold $\tau$, number of semantic categories $C$.
\Ensure Fused global occupancy map $\mathcal{M}_{\text{global}} \in \{0, 1, \dots, C\}^{X \times Y \times Z}$.

\State \texttt{// Phase 1: SE(2) Projection \& Keyframe Selection}
\State $\mathbf{T}_t(x, y) \gets \text{Proj}_{\text{SE(2)}}(\mathbf{T}_t), \quad \forall t \in \{1 \dots T\}$ \Comment{Align to BEV plane}
\State $\mathcal{K} \gets \{1\}, \quad \mathbf{p}_{\text{last}} \gets \mathbf{T}_1(x,y)$
\For{$t = 2$ \textbf{to} $T$}
    \If{$\|\mathbf{T}_t(x,y) - \mathbf{p}_{\text{last}}\|_2 > d_{\max}$}
        \State $\mathcal{K} \gets \mathcal{K} \cup \{t\}, \quad \mathbf{p}_{\text{last}} \gets \mathbf{T}_t(x,y)$
    \EndIf
\EndFor

\State \texttt{// Phase 2: Pass 1 - Keyframe Sinking \& Fusion}

\State Initialize $\mathcal{M}_{\text{global}}(\mathbf{v}) \gets \varnothing, \quad \forall \mathbf{v} \in \mathbb{Z}^{X \times Y \times Z}$ \Comment{$\varnothing$ denotes unassigned}.
\For{$k \in \mathcal{K}$}
    \State $\mathcal{O}_k^{\prime} \gets \mathcal{W}(\hat{\mathcal{O}_k}, \mathbf{T}_k)$ \Comment{Spatial warping to global coordination}
    \State $\triangleright$ Assign key-frame voxels from local map to global map.
    \State $\mathcal{M}_{\text{global}}(\mathbf{v}) \gets \mathcal{O}^{\prime}_k(\mathbf{v}), \quad \forall \mathbf{v} \text{ s.t. } \mathcal{M}_{\text{global}}(\mathbf{v}) = \varnothing$ 
\EndFor

\State $\Delta z(x,y) \gets \min \{ z \mid \mathcal{M}_{\text{global}}(x,y,z) \in \mathcal{C}_{\text{ground}} \}$
\State $\mathcal{M}_{\text{global}}(x,y,z) \gets \mathcal{M}_{\text{global}}(x,y, z + \Delta z)$ \Comment{Column sinking to lowest ground}

\State $\mathcal{M}_{\text{global}}(\mathbf{v}) \gets \text{Mode}(\mathcal{N}_{3\times3}(\mathbf{v}))$ for $\mathbf{v}$ where $\mathcal{M}_{\text{global}}(\mathbf{v}) = \varnothing, z=0$

\State \texttt{// Phase 3: Pass 2 - Non-Keyframe Voting Inpainting}
\State Initialize vote tensor $\mathcal{V} \in \mathbb{N}^{X \times Y \times Z \times C} \gets \mathbf{0}$
\For{$t \notin \mathcal{K}$}
    \State $\mathcal{O}_t^{\prime} \gets \mathcal{W}(\hat{\mathcal{O}}_t, \mathbf{T}_t)$
    \State $\mathcal{V}(\mathbf{v}, c) \gets \mathcal{V}(\mathbf{v}, c) + \mathbb{I}[\mathcal{O}^{\prime}_t(\mathbf{v}) = c], \quad \forall \mathbf{v} \text{ s.t. } \mathcal{M}_{\text{global}}(\mathbf{v}) = \varnothing$
\EndFor
\State $\mathcal{M}_{\text{global}}(\mathbf{v}) \gets \arg\max_c \mathcal{V}(\mathbf{v}, c)$ if $\max_c \mathcal{V}(\mathbf{v}, c) \ge \tau$

\State \texttt{//   Phase 4: Morphological Refinement}
\State $\mathcal{M}_{\text{global}} \gets \text{CloseOp}(\mathcal{M}_{\text{global}}, c_{\text{sidewalk}})$ \Comment{Binary closing on sidewalk}
\State $\mathcal{M}_{\text{global}} \gets \text{FilterAreaOp}(\mathcal{M}_{\text{global}}, c_{\text{road}}, A < 2\text{m}^2)$ \Comment{Remove isolated noise}
\State \Return $\mathcal{M}_{\text{global}}$
\end{algorithmic}
\end{algorithm}

The pipeline consists of four primary phases. First, to manage computational overhead and redundancy, we subsample the generated sequences into keyframes based on a spatial distance threshold, projecting them onto a unified SE(2) bird's-eye-view (BEV) plane. Second, during the initial fusion pass, we introduce a deterministic column sinking operation. This forces all predicted ground-level voxels to align with the absolute zero-elevation plane ($z=0$), effectively rectifying pitch and roll artifacts or elevation drifts accumulated over long-horizon rollouts. Third, to address occlusions and local generation artifacts, we leverage the temporal redundancy of non-keyframes; by applying a spatial majority voting mechanism, we robustly inpaint the remaining gaps on the $z=0$ plane. Finally, a morphological refinement stage—comprising binary closing and connected-component area filtering—is applied to enforce spatial continuity on sidewalks and systematically remove isolated noise clusters (e.g., artifacts smaller than $2\text{m}^2$). This pipeline ensures a connected and smooth road topology, which is strictly required for downstream 2D multi-agent simulations.

\subsection{Branch road detection and topology extraction.} 
\label{sec:topology extraction}
Then, to bridge the gap between discrete occupancy grids $\mathcal{M}_{\text{global}}$ and vectorized road networks required for downstream multi-agent simulation, we propose a robust topology extraction and graph refinement algorithm. We first extract the 2D drivable surface mask, apply connected-component filtering, and perform strict single-pixel skeletonization to initialize the graph structure. To mitigate rasterization artifacts, we systematically eliminate spurious triangular loops by removing the longest edges within 3-cliques. Subsequently, a rigorous graph cleaning protocol is executed: we iteratively prune short, noisy spurs and introduce a node contraction mechanism to merge spatially dense junction clusters into unified super-nodes, accurately reflecting complex real-world intersections. Finally, to precisely isolate valid map ingress and egress points, we design a novel Dual-Probe filtering mechanism. A topology probe identifies and discards artificial endpoints caused by internal fragmentation, while a semantic probe leverages the 3D elevation profile of the occupancy map to detect physical occlusions or obstacles along the projected path. Ultimately, this pipeline translates the generated voxel map into a high-fidelity, deadlock-free topological graph, providing a reliable foundation for downstream closed-loop planning and control.

\begin{algorithm}[h]
\caption{Branch point detection / topology extraction}
\label{alg:branch_detection}
\begin{algorithmic}[1]
\Require Fused global map $\mathcal{M}_{\text{global}} \in \{0, \dots, C\}^{X \times Y \times Z}$, lane width $w_{\text{lane}}$, pruning threshold $\tau_{\text{prune}}$. $\mathcal{V}_{\text{valid}} \gets \varnothing$ for valid branch node set in the graph. 
\Ensure Vectorized road graph $\mathcal{G} = (V, E)$, valid endpoints $\mathbf{P}_{\text{valid}}$.

\State \texttt{// Phase 1: Binary Masking \& Skeletonization}
\State $\mathcal{B}(x,y) \gets \mathbb{I}[\mathcal{M}_{\text{global}}(x, y, 0) = c_{\text{road}}], \quad \forall (x,y) \in \mathbb{Z}^2$
\State $\mathcal{S} \gets \text{Skeletonize}(\mathcal{B})$
\State $\mathcal{S}(x+1, y+1) \gets 0, \quad \forall (x,y) \text{ s.t. } \sum_{i=0}^{1} \sum_{j=0}^{1} \mathcal{S}(x+i, y+j) = 4$ 

\State \texttt{// Phase 2: Graph Initialization \& Artifact Removal}
\State $V \gets \{ \mathbf{v} \in \mathbb{Z}^2 \mid \mathcal{S}(\mathbf{v}) = 1 \}$
\State $E \gets \{ (\mathbf{u}, \mathbf{v}) \mid \mathbf{u}, \mathbf{v} \in V, \|\mathbf{u} - \mathbf{v}\|_{\infty} = 1 \}$, with weight $w(\mathbf{u}, \mathbf{v}) = \|\mathbf{u} - \mathbf{v}\|_2$
\State Initialize graph $\mathcal{G} \gets (V, E)$
\For{each 3-clique (triangle) $K_3 \in \mathcal{G}$} 
    \State $e_{\text{max}} \gets \arg\max_{e \in K_3} w(e)$
    \State $E \gets E \setminus \{e_{\text{max}}\}$ \Comment{Break loops to expose leaf nodes}
\EndFor

\State \texttt{// Phase 3: Iterative Spur Pruning \& Junction Contraction}
\Repeat
    \State $\mathcal{V}_{\text{leaf}} \gets \{v \in V \mid \text{deg}(v) = 1 \}$
    \State Find branch path $P_{l \to j}$ from $l \in \mathcal{V}_{\text{leaf}}$ to nearest junction $j$ ($\text{deg}(j) > 2$)
    \If{$\sum_{e \in P} w(e) < \tau_{\text{prune}}$}
        \State $V \gets V \setminus (P_{l \to j} \setminus \{j\})$ \Comment{Remove short spurs}
    \EndIf
\Until{no spurs are removed}
\State $\mathcal{V}_{\text{junc}} \gets \{v \in V \mid \text{deg}(v) > 2 \}$

\For{$\mathbf{u}, \mathbf{v} \in \mathcal{V}_{\text{junc}}$ s.t. $\|\mathbf{u} - \mathbf{v}\|_2 < 2 \cdot w_{\text{lane}}$} 
    \State $\mathcal{G} \gets \text{ContractNodes}(\mathcal{G}, \mathbf{u}, \mathbf{v})$ \Comment{Merge clustered junctions}
\EndFor

\State \texttt{// Phase 4: Dual-Probe Endpoint Filtering}
\State $\mathcal{O}_{\text{obstacle}} \gets (\mathcal{M}_{\text{global}}(x,y,z \ge 1) \notin \{c_{\text{road}}, c_{\text{sidewalk}}, \varnothing\})$
\Comment{Get 3D obstacle occ above ground layer}

\State $\mathcal{V}_{\text{end}} \gets \{v \in V \mid \text{deg}(v) = 1 \}$
\For{$v \in \mathcal{V}_{\text{end}}$}
    \State Let $\vec{d}$ be the outward normalized direction vector from parent junction to $v$.
    \State \textit{Topology Probe}: $p_{\text{topo}} \gets v + (1.5 \cdot w_{\text{lane}}) \vec{d}$
    \State \textit{Semantic Probe}: Generate $15\text{m} \times 3.6\text{m}$ bounding box $B_{\text{sem}}$ along $\vec{d}$ from $v$.
    \If{$\mathcal{B}(p_{\text{topo}}) == 0$ \textbf{and} $\sum_{\mathbf{p} \in B_{\text{sem}}} \mathbb{I}[\mathcal{O}_{\text{3D}}(\mathbf{p})] < \tau_{\text{obs}}$}
        \State $\mathcal{V}_{\text{valid}} \gets \mathcal{V}_{\text{valid}} \cup \{v\}$ \Comment{Endpoint is neither internal nor blocked}
    \EndIf
\EndFor

\State \Return $\mathcal{G}$, $\mathcal{V}_{\text{valid}}$
\end{algorithmic}
\end{algorithm}

\subsection{Lane topology extraction.} 
\label{sec:lane_topo}

After we have $\mathcal{G}$, $\mathcal{V}_{\text{valid}}$, we use the retained graph information to create lane graphs.  As shown in Alg.~\ref{alg:lane_topology}, based on the extracted skeleton, i.e., $E \in \mathcal{G}$, we introduce a two-phase lane topology extraction algorithm (Algorithm 3). In the first phase, we convert the raw skeletal segments into kinematically smooth reference centerlines using 3rd-degree B-Spline interpolation with a fixed spatial resolution $\Delta s_{\text{step}}$. To reconstruct multi-lane structures, we generate explicit parallel lanes via orthogonal offsetting along the normal vectors of the reference trajectories. To guarantee physical feasibility, we strictly enforce a topological boundary constraint: the interior of any newly generated lane must be fully bounded by the drivable surface mask $\mathcal{B}$.

In the second phase, we address the geometric overlaps that inevitably occur at complex intersections where multiple segments converge. We formulate this as a spatial intersection resolution problem. For each candidate lane, we identify conflict points that fall within a minimal distance threshold $\epsilon$ of any other independent lane. We then sever the trajectories at these overlapping junctions and exclusively retain the longest continuous sub-trajectory. Finally, a secondary B-Spline interpolation is applied to eliminate any geometric artifacts or non-differentiable kinks introduced by the splitting process. This rigorous pipeline yields a highly precise, non-overlapping lane network, providing a kinematically reliable foundation for multi-agent closed-loop simulation. 

\begin{algorithm}[t]
\caption{Lane Topology Extraction}
\label{alg:lane_topology}
\begin{algorithmic}[1]

\Require Extracted segments $\mathcal{S}$ from $\mathcal{G}$, drivable mask $\mathcal{B}$, lane width $w_{\text{lane}}$, distance threshold $\epsilon$, resampling step $\Delta s_{\text{step}}$.
\Ensure Vectorized explicit lanes $\mathcal{L}_{\text{final}}$.

\State \texttt{// Phase 1: B-Spline Smoothing \& Parallel Offsetting}
\State $\mathcal{L}_{\text{cand}} \gets \varnothing$
\For{each $S \in \mathcal{S}$ s.t. $|S| \ge 10$}
    \State $\mathbf{P}_c \gets \text{BSpline}(S_{\text{path}}, \text{deg} = 3, \Delta s_{\text{step}})$
    \State $\mathbf{N} \gets \text{NormalVectors}(\mathbf{P}_c)$, $n \gets \lfloor S_{\text{width}} / w_{\text{lane}} \rfloor - 1$
    \State $\mathbf{O} \gets \left\{ \left(i - \frac{n-1}{2}\right) w_{\text{lane}} \;\middle|\; i \in \{0, \dots, n-1\} \right\}$
    \For{$o \in \mathbf{O}$}
        \State $\mathbf{P}_l \gets \mathbf{P}_c + o \cdot \mathbf{N}$ \Comment{Generate parallel lane curves}
        \If{$\mathbf{P}_l^{\circ} \subset \mathcal{B}$} \Comment{$\mathbf{P}_l^{\circ}$ denotes the interior of the trajectory}
            \State $\mathcal{L}_{\text{cand}} \gets \mathcal{L}_{\text{cand}} \cup \{\mathbf{P}_l\}$
        \EndIf
    \EndFor
\EndFor

\State \texttt{// Phase 2: Spatial Intersection Resolution \& Resmoothing}
\For{each $L \in \mathcal{L}_{\text{cand}}$}
    \State $\mathcal{C}_L \gets \{ \mathbf{p} \in L \mid \exists L' \in \mathcal{L}_{\text{cand}} \setminus \{L\}, \text{dist}(\mathbf{p}, L') < \epsilon \}$ \Comment{Identify cross-lane overlaps}
    \State $L \gets \arg\max_{L_{\text{sub}} \subseteq (L \setminus \mathcal{C}_L)} |L_{\text{sub}}|$ \Comment{Retain the longest continuous sub-trajectory}
    \State $L \gets \text{BSpline}(L, \text{deg} = 3, \Delta s_{\text{step}})$ \Comment{Smooth artifacts caused by splitting}
\EndFor
\State $\mathcal{L}_{\text{final}} \gets \{ L \in \mathcal{L}_{\text{cand}} \mid |L| \ge 5 \}$ \Comment{Remove all lanes short than 5 pixels}

\State \Return $\mathcal{L}_{\text{final}}$

\end{algorithmic}
\end{algorithm}

\begin{algorithm}[tb]
    \caption{Spatially-conditioned agent generation and routing initialization}
    \label{alg:spawnagents}
    \begin{algorithmic}[1]

    \Require Query anchor pose $\mathbf{T} \in SE(2)$ with spatial coordinate $\mathbf{p}_{\mathbf{T}}$, boolean flag $b_{\text{ego}}$ indicating whether to explicitly initialize the ego vehicle at $\mathbf{T}$, mean/standard of initial velocity $\mu_v, \sigma_v$, $\mathcal{M}_{\text{global}}$ as global map, $\mathcal{V}_{\text{valid}}$ from Alg.~\ref{alg:branch_detection}, Agents asset pool $\mathcal{X}$.
    
    \State Crop local occupancy $\mathcal{O}_{\mathbf{T}}$ from $\mathcal{M}_{\text{global}}$ by pose $\mathbf{T}$.
    \State $\mathcal{A}_{\text{new}} \gets \text{LayoutGenerator}(\mathcal{O}_{\mathbf{T}}) \cup \big( \{\mathbf{p}_{\mathbf{T}}\} \text{ if } b_{\text{ego}} \text{ else } \emptyset \big)$ 

    \For{each agent $a \in \mathcal{A}_{\text{new}}$ parameterized by pose $\mathbf{p}_a$}
        \State Initialize velocity $v_a \gets v_{\text{des}} \sim \mathcal{N}(\mu_v, \sigma_v)$
        \State $\mathbf{p}_{\text{target}} \sim \text{Uniform}(\mathcal{V}_{\text{valid}})$  \Comment{Random valid endpoint}
        \State $\mathcal{x}_a \sim \text{Uniform}(\mathcal{X})$ \Comment{Random agent asset}
        
        \State $\pi_{\text{traj}} \gets \text{A*}(\mathcal{L}_{\text{final}}, \mathbf{p}_a, \mathbf{p}_{\text{target}})$ \Comment{Shortest valid path}
        \State $\mathbf{h}_a \gets \text{Tangent}(\pi_{\text{traj}}, \mathbf{p}_a)$ \Comment{Generate initial headings from trajectory}
        \State $a \gets (\mathbf{p}_a, \mathbf{h}_a, v_a, \pi_{\text{traj}}, \mathcal{x}_a)$ \Comment{Assign initial states to agent}
    \EndFor
    \State \Return $\mathcal{A}_{\text{new}}$
    \end{algorithmic}
\end{algorithm}

\subsection{2D-IDM based agent simulation engine.} 
\label{sec:2d-idm}

With all geometric information obtained, we introduce a 2D-IDM based simulation in Alg.~\ref{alg:idm_rollout}. The algorithm is structurally divided into two phases: global pre-computation (Phase 1) and dynamic step-by-step execution (Phase 2).

Phase 1 employs a proactive spatial pre-computation strategy. Rather than solely initializing traffic around the starting ego-pose ($\mathbf{T}_0^{\text{ego}}$), the engine seamlessly injects background agents at forward and backward anchor poses ($\mathbf{T}_{\text{fwd}}$, $\mathbf{T}_{\text{bwd}}$) extending up to a predefined horizon $d_{\text{pre}}$. The unified SpawnAgents procedure handles this layout generation, simultaneously sampling valid endpoints, predicting spatial coordinates, and assigning initial shortest paths via A* routing.

In the simulation part, \cref{alg:idm_rollout} Line~\ref{algline:leading_vehicles}
defines the spatial identification of candidate leading vehicles ($\mathcal{C}_a$). Instead of relying on heuristic bounding box checks, an agent $a$ filters surrounding vehicles using a normalized vector dot product, $\bigg(\frac{(\mathbf{p}_{a'} - \mathbf{p}_a) \cdot \mathbf{h}_a}{\|\mathbf{p}_{a'} - \mathbf{p}_a\|_2} > 0.5\bigg)$, to ensure the target is strictly within a $60^\circ$ forward field-of-view cone. This is coupled with a trajectory distance constraint ($\text{Dist}(\mathbf{p}_{a'}, \pi_{\text{traj}}) < d_{\text{lat}}$) to confirm the target is actively blocking the planned route.

Furthermore, \cref{alg:idm_rollout} Line~\ref{algline:lanechange_logic}
explicitly shows the lane-changing logic. An agent initiates a Bezier-smoothed lane transition and A* re-routing only when the longitudinal distance falls below a critical threshold ($s < d_{\text{lc}}$) under two specific risks: either the ego agent is approaching a slower leader ($\Delta v > 0$), or it encounters a wrong-way/head-on vehicle. The latter scenario is elegantly captured by the negative dot product of their respective heading unit vectors ($\mathbf{h}_a \cdot \mathbf{h}_{a_{\text{lead}}} < -0.5$).

Finally, the kinematic states are updated via IDM, and the dynamic 3D assets are sequentially rendered onto the cropped static background using a spatial overwrite operator ($\circledast$) to handle semantic Z-buffering (Step 2.3), ultimately producing a seamless sequence of dynamic local environments.

\begin{algorithm}[htb]
\caption{A* routing + 2D-IDM based control engine pseudo codes}
\label{alg:idm_rollout}
\begin{algorithmic}[1]

\Require A pool of global maps $\mathbb{M}$ generated, IDM parameters, rolling horizon threshold $d_{\text{roll}}$, simulation horizon $\phi$, Lane topology $\mathcal{L}_{\text{final}}$, $\Delta t$ for Inter-frame time interval, $\text{FOV}(\mathbf{T})$: The set of all spatial coordinates within the perception range of the ego vehicle at pose $\mathbf{T}$, Agents asset pool $\mathcal{X}$, $d_{\text{lc}}$ for distance threshold of lane change. SpawnAgents: method generate initial state of agents, details in Alg.~\ref{alg:spawnagents}.

\Ensure Rendered sequence of dynamic local environments $\{\mathcal{M}_{\text{local}}^{(t)}\}_{t=1}^{\phi}$.

\State \texttt{// Phase 1: Global Initialization \& Pre-computation}
\State Randomly select a global map $\mathcal{M}_{\text{global}} \sim \text{Uniform}(\mathbb{M})$.
\State Get ego-poses $\{\mathbf{T}_t\}_{t=1}^T$ used to build $\mathcal{M}_{\text{global}}$ in Alg.~\ref{alg:keyframe_fusion}.
\State Randomly select initial ego pose $\mathbf{T}_0^{\text{ego}} \sim \text{Uniform}(\{\mathbf{T}_t\}_{t=1}^T)$.
\State Find $\mathbf{T}_{\text{fwd}}$ and $\mathbf{T}_{\text{bwd}}$ at distance $\pm d_{\text{pre}}$ along $\{\mathbf{T}_t\}_{t=1}^T$ from $\mathbf{T}_0^{\text{ego}}$.
\State $\mathcal{A} \gets \text{SpawnAgents}(\mathbf{T}_0^{\text{ego}}, \text{True}) \cup \text{SpawnAgents}(\mathbf{T}_{\text{fwd}}) \cup \text{SpawnAgents}(\mathbf{T}_{\text{bwd}})$

\State \texttt{// Phase 2: Closed-Loop Simulation}
\State $\Delta d_{\text{ego}} \gets 0, \quad \mathcal{S}_{\text{out}} \gets \emptyset$ \Comment{Init movement tracker and empty sequence for frames}
\For{$t = 1 \dots \phi$}
    \State \texttt{// Step 2.1: Rolling Horizon Update}
    \State $\Delta d_{\text{ego}} \gets \Delta d_{\text{ego}} + v_{\text{ego}}^{(t)} \Delta t$
    \If{$\Delta d_{\text{ego}} \ge d_{\text{roll}}$}
        \State $\mathcal{A} \gets \{a \in \mathcal{A} \mid \mathbf{p}_a \in \text{FOV}(\mathbf{T}_t^{\text{ego}})\}$ \Comment{Remove out-of-view agents}
        \State $\mathbf{T}_{\text{fwd}} \gets \text{Pose ahead of } \mathbf{T}_t^{\text{ego}} \text{ by } d_{\text{pre}}$, $\mathbf{T}_{\text{bwd}} \gets \text{Pose behind of } \mathbf{T}_t^{\text{ego}} \text{ by } d_{\text{pre}}$
        \State $\mathcal{A} \gets \mathcal{A} \cup \text{SpawnAgents}(\mathbf{T}_{\text{fwd}}) \cup \text{SpawnAgents}(\mathbf{T}_{\text{bwd}})$ 
        \State $\Delta d_{\text{ego}} \gets 0$
    \EndIf

\State \texttt{// Step 2.2: IDM Kinematics \& Route Execution}
    \For{each agent $a \in \mathcal{A}$ parameterized by state $(\mathbf{p}_a,  \mathbf{h}_a, v_a, \pi_{\text{traj}}, \mathbf{p}_{\text{target}})$}
    \State Let $\mathbf{h}_a$ denote the heading unit vector of agent $a$.
    
    \State $ \mathcal{C}_a \gets \left\{ a' \in \mathcal{A} \setminus \{a\} \;\middle|\; \frac{(\mathbf{p}_{a'} - \mathbf{p}_a) \cdot \mathbf{h}_a}{\|\mathbf{p}_{a'} - \mathbf{p}_a\|_2} > 0.5 \land \text{Dist}(\mathbf{p}_{a'}, \pi_{\text{traj}}) < d_{\text{lat}} \right\}$ \label{algline:leading_vehicles}
        
    \State $a_{\text{lead}} \gets \arg\min_{a' \in \mathcal{C}_a} \|\mathbf{p}_{a'} - \mathbf{p}_a\|_2$ \Comment{Select out the agent ahead of current a}

    \State $s \gets \|\mathbf{p}_{a_{\text{lead}}} - \mathbf{p}_a\|_2$, $\Delta v \gets v_a - v_{\text{lead}}$ \Comment{adjusted if head-on}
    
    \If{$s < d_{\text{lc}}$ \textbf{and} $(\Delta v > 0 \textbf{ or } \mathbf{h}_a \cdot \mathbf{h}_{a_{\text{lead}}} < -0.5)$} \label{algline:lanechange_logic}
        \State $\mathbf{p}_{\text{adj}} \gets \text{SearchParallelLane}(\mathbf{p}_a, \mathcal{L}_{\text{final}})$
        \If{$\mathbf{p}_{\text{adj}} \neq \emptyset$}
            \State $\pi_{\text{bezier}} \gets \text{BezierCurve}(\mathbf{p}_a, \mathbf{p}_{\text{adj}})$ \Comment{Generate smooth transition}
            \State $\pi_{\text{new}} \gets \text{A-Star}(\mathcal{L}_{\text{final}}, \mathbf{p}_{\text{adj}}, \mathbf{p}_{\text{target}})$ \Comment{Re-route from adjacent lane}
            \State $\pi_{\text{traj}} \gets \pi_{\text{bezier}} \oplus \pi_{\text{new}}$ \Comment{Concatenate trajectories}
            \State $a_{\text{acc}} \gets \text{IDM}(v_a, v_{\text{des}}, \Delta v, s)$ \hfill \Comment{Compute acceleration}
            \State $\mathbf{p}_a \gets \text{Advance}(\mathbf{p}_a, \pi_{\text{traj}}, v_a \Delta t)$ \Comment{Move along trajectory}
        \EndIf
    \EndIf
    \EndFor
            
\State \texttt{// Step 2.3: Rendering}
    \State $\mathcal{A}_{\text{vis}} \gets \{a \in \mathcal{A} \mid \mathbf{p}_a \in \text{FOV}(\mathbf{T}_t^{\text{ego}})\}$ \hfill \Comment{Filter visible agents in current view}
    \State $\mathcal{O}_{\mathbf{T}}^{(t)} \gets \text{Crop}(\mathcal{M}_{\text{global}}, \mathbf{T}_t^{\text{ego}})$ \hfill \Comment{Extract static local background}

    \State $\mathcal{X}^{(t)} \gets \bigcup_{a \in \mathcal{A}_{\text{vis}}} \Phi_{\text{rigid}}(\S_a, \mathbf{p}_a, \mathbf{h}_a)$ \hfill \Comment{Transform and aggregate dynamic assets}
    \State $\mathcal{M}_{\text{local}}^{(t)} \gets \mathcal{O}_{\mathbf{T}}^{(t)} \circledast \mathcal{X}^{(t)}$ \hfill \Comment{Overlay foreground onto static background}
    \State $\mathcal{S}_{\text{out}} \gets \mathcal{S}_{\text{out}} \cup \{\mathcal{M}_{\text{local}}^{(t)}\}$ \Comment{Append rendered frame to sequence}

    \EndFor

\State \Return $\mathcal{S}_{\text{out}}$

\end{algorithmic}
\end{algorithm}

\newpage
\section{Potential limitation and Future work}
\label{sec:limitation}
Although OccSim provide the first implementation of using occupancy world model in simulation, we still observe the following limitation in design and evaluation period of OccSim:

\begin{itemize}
    \item \textbf{Heuristic based fusion and lane graph extraction}: Although the robustness of our multi-frame fusion/lane graph extraction algorithm has been qualitatively demonstrated in numerous practical applications, given the occasional instability in the quality of the raw occupancy frames output by the model, we still seek an end-to-end fusion approach rather than frame-by-frame fusion.
    
    \item \textbf{Latent-space warping.} W-DiT warps in latent space, assuming the VAE is approximately equivariant to planar rigid transforms. The U-Net refinement absorbs the resulting interpolation error in practice, but warping directly in occupancy space or learning an equivariant latent would remove this approximation.
    
    \item \textbf{Limited training data}: Compared to image data, semantic occupancy data requires manual semantic annotation, which limits the total amount of available training data to fewer than 100,000 frames. This volume is significantly smaller than the amount of available RGB video data, and it also restricts our ability to explore the upper limits of the model’s performance. 

    \item \textbf{IDM based agent behavior simulation}: In this paper, we use IDM, a classic traffic flow control model to simulate the behavior of vehicles. Although this approach has proven effective in downstream tasks, we still hope to find more effective ways to enable this aspect of control to learn from the data, or even integrate it into the process of generating static occupancy.

    \item \textbf{Open-loop evaluation.} Existing closed-loop driving benchmarks expose either raw sensor inputs (e.g., NAVSIM) or vectorized map and object states (e.g., nuPlan, Waymax), and provide no semantic occupancy interface, so occupancy-input models cannot be evaluated there in closed loop. All our downstream metrics (mIoU, IoU and planning L2) are therefore computed open-loop against logged futures; closed-loop evaluation of learned policies inside OccSim is left for future work.
    
\end{itemize}

\clearpage\clearpage
\bibliographystyle{plainnat}
\bibliography{main}

@String(ECCV  = {Eur. Conf. Comput. Vis.})

@String(ECCV  = {ECCV})

@inproceedings{Dosovitskiy17,
  title={CARLA: An Open Urban Driving Simulator},
  author={Alexey Dosovitskiy and German Ros and Felipe Codevilla and Antonio Lopez and Vladlen Koltun},
  booktitle={Proceedings of the 1st Annual Conference on Robot Learning},
  pages={1--16},
  year={2017},
  editor={A. D. Dragan and M. Hebert and S. Levine},
  volume={78},
  series={Proceedings of Machine Learning Research},
  address={Mountain View, CA, USA},
  month={13--15 Nov},
  publisher={PMLR},
  url={http://proceedings.mlr.press/v78/dosovitskiy17a.html}
}

@inproceedings{chitta2024sledge,
  title={Sledge: Synthesizing driving environments with generative models and rule-based traffic},
  author={Chitta, Kashyap and Dauner, Daniel and Geiger, Andreas},
  booktitle={European Conference on Computer Vision},
  pages={57--74},
  year={2024},
  organization={Springer}
}

@article{liu2025towards,
  title={Towards foundational LiDAR world models with efficient latent flow matching},
  author={Liu, Tianran and Zhao, Shengwen and Rhinehart, Nicholas},
  journal={arXiv preprint arXiv:2506.23434},
  year={2025}
}

@inproceedings{lugmayr2022repaint,
  title={Repaint: Inpainting using denoising diffusion probabilistic models},
  author={Lugmayr, Andreas and Danelljan, Martin and Romero, Andres and Yu, Fisher and Timofte, Radu and Van Gool, Luc},
  booktitle={Proceedings of the IEEE/CVF conference on computer vision and pattern recognition},
  pages={11461--11471},
  year={2022}
}

@article{gao2024vista,
  title={Vista: A generalizable driving world model with high fidelity and versatile controllability},
  author={Gao, Shenyuan and Yang, Jiazhi and Chen, Li and Chitta, Kashyap and Qiu, Yihang and Geiger, Andreas and Zhang, Jun and Li, Hongyang},
  journal={Advances in Neural Information Processing Systems},
  volume={37},
  pages={91560--91596},
  year={2024}
}

@article{zhang2025mem2ego,
  title={Mem2ego: Empowering vision-language models with global-to-ego memory for long-horizon embodied navigation},
  author={Zhang, Lingfeng and Liu, Yuecheng and Zhang, Zhanguang and Aghaei, Matin and Hu, Yaochen and Gu, Hongjian and Alomrani, Mohammad Ali and Bravo, David Gamaliel Arcos and Karimi, Raika and Hamidizadeh, Atia and others},
  journal={arXiv preprint arXiv:2502.14254},
  year={2025}
}

@article{song2025hume,
  title={Hume: Introducing system-2 thinking in visual-language-action model},
  author={Song, Haoming and Qu, Delin and Yao, Yuanqi and Chen, Qizhi and Lv, Qi and Tang, Yiwen and Shi, Modi and Ren, Guanghui and Yao, Maoqing and Zhao, Bin and others},
  journal={arXiv preprint arXiv:2505.21432},
  year={2025}
}

@article{xiao2025splatco,
  title={SplatCo: Structure-View Collaborative Gaussian Splatting for Detail-Preserving Rendering of Large-Scale Unbounded Scenes},
  author={Xiao, Haihong and Zou, Jianan and Zhou, Yuxin and He, Ying and Kang, Wenxiong},
  journal={arXiv preprint arXiv:2505.17951},
  year={2025}
}

@article{chen2024lidar,
  title={Lidar-gs: Real-time lidar re-simulation using gaussian splatting},
  author={Chen, Qifeng and Yang, Sheng and Du, Sicong and Tang, Tao and Xie, Rengan and Chen, Peng and Huo, Yuchi},
  journal={arXiv preprint arXiv:2410.05111},
  year={2024}
}

@article{hu2024drivingworld,
  title={DrivingWorld: Constructing world model for autonomous driving via video GPT},
  author={Hu, Xiaotao and Yin, Wei and Jia, Mingkai and Deng, Junyuan and Guo, Xiaoyang and Zhang, Qian and Long, Xiaoxiao and Tan, Ping},
  journal={arXiv preprint arXiv:2412.19505},
  year={2024}
}

@article{zhang2025epona,
  title={Epona: Autoregressive Diffusion World Model for Autonomous Driving},
  author={Zhang, Kaiwen and Tang, Zhenyu and Hu, Xiaotao and Pan, Xingang and Guo, Xiaoyang and Liu, Yuan and Huang, Jingwei and Yuan, Li and Zhang, Qian and Long, Xiao-Xiao and others},
  journal={arXiv preprint arXiv:2506.24113},
  year={2025}
}

@article{agarwal2025cosmos,
  title={Cosmos world foundation model platform for physical ai},
  author={Agarwal, Niket and Ali, Arslan and Bala, Maciej and Balaji, Yogesh and Barker, Erik and Cai, Tiffany and Chattopadhyay, Prithvijit and Chen, Yongxin and Cui, Yin and Ding, Yifan and others},
  journal={arXiv preprint arXiv:2501.03575},
  year={2025}
}

@article{ho2022classifier,
  title={Classifier-free diffusion guidance},
  author={Ho, Jonathan and Salimans, Tim},
  journal={arXiv preprint arXiv:2207.12598},
  year={2022}
}

@article{ren2025cosmos,
  title={Cosmos-Drive-Dreams: Scalable Synthetic Driving Data Generation with World Foundation Models},
  author={Ren, Xuanchi and Lu, Yifan and Cao, Tianshi and Gao, Ruiyuan and Huang, Shengyu and Sabour, Amirmojtaba and Shen, Tianchang and Pfaff, Tobias and Wu, Jay Zhangjie and Chen, Runjian and others},
  journal={arXiv preprint arXiv:2506.09042},
  year={2025}
}

@article{hu2023gaia,
  title={Gaia-1: A generative world model for autonomous driving},
  author={Hu, Anthony and Russell, Lloyd and Yeo, Hudson and Murez, Zak and Fedoseev, George and Kendall, Alex and Shotton, Jamie and Corrado, Gianluca},
  journal={arXiv preprint arXiv:2309.17080},
  year={2023}
}

@article{ma2024latte,
  title={Latte: Latent diffusion transformer for video generation},
  author={Ma, Xin and Wang, Yaohui and Chen, Xinyuan and Jia, Gengyun and Liu, Ziwei and Li, Yuan-Fang and Chen, Cunjian and Qiao, Yu},
  journal={arXiv preprint arXiv:2401.03048},
  year={2024}
}

@article{russell2025gaia,
  title={Gaia-2: A controllable multi-view generative world model for autonomous driving},
  author={Russell, Lloyd and Hu, Anthony and Bertoni, Lorenzo and Fedoseev, George and Shotton, Jamie and Arani, Elahe and Corrado, Gianluca},
  journal={arXiv preprint arXiv:2503.20523},
  year={2025}
}

@inproceedings{li2025uniscene,
  title={Uniscene: Unified occupancy-centric driving scene generation},
  author={Li, Bohan and Guo, Jiazhe and Liu, Hongsi and Zou, Yingshuang and Ding, Yikang and Chen, Xiwu and Zhu, Hu and Tan, Feiyang and Zhang, Chi and Wang, Tiancai and others},
  booktitle={Proceedings of the Computer Vision and Pattern Recognition Conference},
  pages={11971--11981},
  year={2025}
}

@article{friedman2022vendi,
  title={The vendi score: A diversity evaluation metric for machine learning},
  author={Friedman, Dan and Dieng, Adji Bousso},
  journal={arXiv preprint arXiv:2210.02410},
  year={2022}
}

@inproceedings{yang2023unisim,
  title={Unisim: A neural closed-loop sensor simulator},
  author={Yang, Ze and Chen, Yun and Wang, Jingkang and Manivasagam, Sivabalan and Ma, Wei-Chiu and Yang, Anqi Joyce and Urtasun, Raquel},
  booktitle={Proceedings of the IEEE/CVF Conference on Computer Vision and Pattern Recognition},
  pages={1389--1399},
  year={2023}
}

@article{yang2025resim,
  title={ReSim: Reliable World Simulation for Autonomous Driving},
  author={Yang, Jiazhi and Chitta, Kashyap and Gao, Shenyuan and Chen, Long and Shao, Yuqian and Jia, Xiaosong and Li, Hongyang and Geiger, Andreas and Yue, Xiangyu and Chen, Li},
  journal={arXiv preprint arXiv:2506.09981},
  year={2025}
}

@inproceedings{yang2025x,
  title={X-scene: Large-scale driving scene generation with high fidelity and flexible controllability},
  author={Yang, Yu and Liang, Alan and Mei, Jianbiao and Ma, Yukai and Liu, Yong and Lee, Gim Hee},
  booktitle={The Thirty-ninth Annual Conference on Neural Information Processing Systems},
  year={2025}
}

@inproceedings{peebles2023scalable,
  title={Scalable diffusion models with transformers},
  author={Peebles, William and Xie, Saining},
  booktitle={Proceedings of the IEEE/CVF international conference on computer vision},
  pages={4195--4205},
  year={2023}
}

@inproceedings{zhang2023adding,
  title={Adding conditional control to text-to-image diffusion models},
  author={Zhang, Lvmin and Rao, Anyi and Agrawala, Maneesh},
  booktitle={Proceedings of the IEEE/CVF international conference on computer vision},
  pages={3836--3847},
  year={2023}
}

@article{van2017neural,
  title={Neural discrete representation learning},
  author={Van Den Oord, Aaron and Vinyals, Oriol and others},
  journal={Advances in neural information processing systems},
  volume={30},
  year={2017}
}

@article{bian2024dynamiccity,
  title={Dynamiccity: Large-scale 4d occupancy generation from dynamic scenes},
  author={Bian, Hengwei and Kong, Lingdong and Xie, Haozhe and Pan, Liang and Qiao, Yu and Liu, Ziwei},
  journal={arXiv preprint arXiv:2410.18084},
  year={2024}
}

@article{shi2025come,
  title={COME: Adding Scene-Centric Forecasting Control to Occupancy World Model},
  author={Shi, Yining and Jiang, Kun and Meng, Qiang and Wang, Ke and Wang, Jiabao and Sun, Wenchao and Wen, Tuopu and Yang, Mengmeng and Yang, Diange},
  journal={arXiv preprint arXiv:2506.13260},
  year={2025}
}

@article{gu2024dome,
  title={Dome: Taming diffusion model into high-fidelity controllable occupancy world model},
  author={Gu, Songen and Yin, Wei and Jin, Bu and Guo, Xiaoyang and Wang, Junming and Li, Haodong and Zhang, Qian and Long, Xiaoxiao},
  journal={arXiv preprint arXiv:2410.10429},
  year={2024}
}

@article{wei2024occllama,
  title={Occllama: An occupancy-language-action generative world model for autonomous driving},
  author={Wei, Julong and Yuan, Shanshuai and Li, Pengfei and Hu, Qingda and Gan, Zhongxue and Ding, Wenchao},
  journal={arXiv preprint arXiv:2409.03272},
  year={2024}
}

@inproceedings{liao2025i2,
  title={I2-World: Intra-inter tokenization for efficient dynamic 4D scene forecasting},
  author={Liao, Zhimin and Wei, Ping and Zhang, Ruijie and Chen, Shuaijia and Wang, Haoxuan and Ren, Ziyang},
  booktitle={Proceedings of the IEEE/CVF International Conference on Computer Vision},
  pages={25810--25819},
  year={2025}
}

@inproceedings{zheng2024occworld,
  title={Occworld: Learning a 3d occupancy world model for autonomous driving},
  author={Zheng, Wenzhao and Chen, Weiliang and Huang, Yuanhui and Zhang, Borui and Duan, Yueqi and Lu, Jiwen},
  booktitle={European conference on computer vision},
  pages={55--72},
  year={2024},
  organization={Springer}
}

@article{tian2023occ3d,
  title={Occ3d: A large-scale 3d occupancy prediction benchmark for autonomous driving},
  author={Tian, Xiaoyu and Jiang, Tao and Yun, Longfei and Mao, Yucheng and Yang, Huitong and Wang, Yue and Wang, Yilun and Zhao, Hang},
  journal={Advances in Neural Information Processing Systems},
  volume={36},
  pages={64318--64330},
  year={2023}
}

@inproceedings{codevilla2018offline,
  title={On offline evaluation of vision-based driving models},
  author={Codevilla, Felipe and Lopez, Antonio M and Koltun, Vladlen and Dosovitskiy, Alexey},
  booktitle={Proceedings of the European conference on computer vision (ECCV)},
  pages={236--251},
  year={2018}
}

@article{caesar2021nuplan,
  title={nuplan: A closed-loop ml-based planning benchmark for autonomous vehicles},
  author={Caesar, Holger and Kabzan, Juraj and Tan, Kok Seang and Fong, Whye Kit and Wolff, Eric and Lang, Alex and Fletcher, Luke and Beijbom, Oscar and Omari, Sammy},
  journal={arXiv preprint arXiv:2106.11810},
  year={2021}
}

@article{dauner2024navsim,
  title={Navsim: Data-driven non-reactive autonomous vehicle simulation and benchmarking},
  author={Dauner, Daniel and Hallgarten, Marcel and Li, Tianyu and Weng, Xinshuo and Huang, Zhiyu and Yang, Zetong and Li, Hongyang and Gilitschenski, Igor and Ivanovic, Boris and Pavone, Marco and others},
  journal={Advances in Neural Information Processing Systems},
  volume={37},
  pages={28706--28719},
  year={2024}
}

@inproceedings{wang2025uniocc,
  title={Uniocc: A unified benchmark for occupancy forecasting and prediction in autonomous driving},
  author={Wang, Yuping and Huang, Xiangyu and Sun, Xiaokang and Yan, Mingxuan and Xing, Shuo and Tu, Zhengzhong and Li, Jiachen},
  booktitle={Proceedings of the IEEE/CVF International Conference on Computer Vision},
  pages={25560--25570},
  year={2025}
}

@inproceedings{he2022masked,
  title={Masked autoencoders are scalable vision learners},
  author={He, Kaiming and Chen, Xinlei and Xie, Saining and Li, Yanghao and Doll{\'a}r, Piotr and Girshick, Ross},
  booktitle={Proceedings of the IEEE/CVF conference on computer vision and pattern recognition},
  pages={16000--16009},
  year={2022}
}

@article{zhou2024hugsim,
  title={Hugsim: A real-time, photo-realistic and closed-loop simulator for autonomous driving},
  author={Zhou, Hongyu and Lin, Longzhong and Wang, Jiabao and Lu, Yichong and Bai, Dongfeng and Liu, Bingbing and Wang, Yue and Geiger, Andreas and Liao, Yiyi},
  journal={arXiv preprint arXiv:2412.01718},
  year={2024}
}

@article{xiong2026unidrive,
  title={UniDrive-WM: Unified Understanding, Planning and Generation World Model For Autonomous Driving},
  author={Xiong, Zhexiao and Ye, Xin and Yaman, Burhan and Cheng, Sheng and Lu, Yiren and Luo, Jingru and Jacobs, Nathan and Ren, Liu},
  journal={arXiv preprint arXiv:2601.04453},
  year={2026}
}

@inproceedings{zhou2025hermes,
  title={Hermes: A unified self-driving world model for simultaneous 3d scene understanding and generation},
  author={Zhou, Xin and Liang, Dingkang and Tu, Sifan and Chen, Xiwu and Ding, Yikang and Zhang, Dingyuan and Tan, Feiyang and Zhao, Hengshuang and Bai, Xiang},
  booktitle={Proceedings of the IEEE/CVF International Conference on Computer Vision},
  pages={27817--27827},
  year={2025}
}

@article{vinitsky2022nocturne,
  title={Nocturne: a scalable driving benchmark for bringing multi-agent learning one step closer to the real world},
  author={Vinitsky, Eugene and Lichtl{\'e}, Nathan and Yang, Xiaomeng and Amos, Brandon and Foerster, Jakob},
  journal={Advances in Neural Information Processing Systems},
  volume={35},
  pages={3962--3974},
  year={2022}
}

@inproceedings{yang2025drivearena,
  title={Drivearena: A closed-loop generative simulation platform for autonomous driving},
  author={Yang, Xuemeng and Wen, Licheng and Wei, Tiantian and Ma, Yukai and Mei, Jianbiao and Li, Xin and Lei, Wenjie and Fu, Daocheng and Cai, Pinlong and Dou, Min and others},
  booktitle={Proceedings of the IEEE/CVF International Conference on Computer Vision},
  pages={26933--26943},
  year={2025}
}

@inproceedings{rowe2025scenario,
  title={Scenario dreamer: Vectorized latent diffusion for generating driving simulation environments},
  author={Rowe, Luke and Girgis, Roger and Gosselin, Anthony and Paull, Liam and Pal, Christopher and Heide, Felix},
  booktitle={Proceedings of the IEEE/CVF Conference on Computer Vision and Pattern Recognition},
  pages={17207--17218},
  year={2025}
}

@inproceedings{yan2025drivingsphere,
  title={Drivingsphere: Building a high-fidelity 4d world for closed-loop simulation},
  author={Yan, Tianyi and Wu, Dongming and Han, Wencheng and Jiang, Junpeng and Zhou, Xia and Zhan, Kun and Xu, Cheng-zhong and Shen, Jianbing},
  booktitle={Proceedings of the Computer Vision and Pattern Recognition Conference},
  pages={27531--27541},
  year={2025}
}

@inproceedings{lu2025infinicube,
  title={Infinicube: Unbounded and controllable dynamic 3d driving scene generation with world-guided video models},
  author={Lu, Yifan and Ren, Xuanchi and Yang, Jiawei and Shen, Tianchang and Wu, Zhangjie and Gao, Jun and Wang, Yue and Chen, Siheng and Chen, Mike and Fidler, Sanja and others},
  booktitle={Proceedings of the IEEE/CVF International Conference on Computer Vision},
  pages={27272--27283},
  year={2025}
}

@inproceedings{lee2024semcity,
    title={{SemCity}: Semantic Scene Generation with Triplane Diffusion},
    author={Lee, Jumin and Lee, Sebin and Jo, Changho and Im, Woobin and Seon, Juhyeong and Yoon, Sung-Eui},
    booktitle={IEEE/CVF Conference on Computer Vision and Pattern Recognition},
    pages = {28337-28347},
    year={2024}
}

@article{kesting2010enhanced,
  title={Enhanced intelligent driver model to access the impact of driving strategies on traffic capacity},
  author={Kesting, Arne and Treiber, Martin and Helbing, Dirk},
  journal={Philosophical Transactions of the Royal Society A: Mathematical, Physical and Engineering Sciences},
  volume={368},
  number={1928},
  pages={4585--4605},
  year={2010},
  publisher={The Royal Society}
}

\end{document}